\pdfoutput=1 
\PassOptionsToPackage{square,sort,comma,numbers}{natbib}  
\documentclass{article}

\usepackage[preprint]{neurips_2023}

\usepackage{enumitem}
\usepackage[utf8]{inputenc} 
\usepackage[T1]{fontenc}    
\usepackage{url}            
\usepackage{booktabs}       
\usepackage{amsfonts}       
\usepackage{nicefrac}       
\usepackage{microtype}      
\usepackage{xcolor}         

\usepackage{bbding}
\usepackage{pifont}
\usepackage{amsmath}  
\usepackage{multirow} 
\usepackage{graphicx}
\usepackage{wrapfig}

\usepackage[utf8]{inputenc} 
\usepackage[T1]{fontenc}    

\usepackage{url}            
\usepackage{booktabs}       
\usepackage{amsfonts}       
\usepackage{nicefrac}       
\usepackage{microtype}      
\usepackage{algpseudocode}  
\usepackage{amsmath}  
\usepackage{multirow} 
\usepackage{graphicx}
\usepackage{epstopdf}
\usepackage{appendix}

\usepackage{subcaption}	
\usepackage[ruled,vlined]{algorithm2e}

\usepackage{hyperref}       
\hypersetup{
    colorlinks=true,
    linkcolor=blue,
    citecolor=green,
    urlcolor=green
}

\def\etal{\emph{et al.}}

\title{Layer-adaptive Structured Pruning Guided by Latency}
\author{
Siyuan Pan$^{1}$\thanks{ Equal contribution. Corresponding to pansiyuan@kuaishou.com} \quad
Linna Zhang$^{1*}$ \quad  Jie Zhang$^{1}$ \quad Xiaoshuang Li$^{2}$ \quad Liang Hou$^{1}$ \quad Xiaobing Tu$^{1}$ \thanks{Corresponding author.} \\
$^1$Kuaishou \qquad $^2$ Shanghai Jiao Tong University}

\begin{document}

\maketitle

\begin{abstract}
Structured pruning can simplify network architecture and improve inference speed. Combined with the underlying hardware and inference engine in which the final model is deployed, better results can be obtained by using latency collaborative loss function to guide network pruning together.
Existing pruning methods that optimize latency have demonstrated leading performance, however, they often overlook the hardware features and connection in the network. To address this problem, we propose a global importance score SP-LAMP(Structured Pruning Layer-Adaptive Magnitude-based Pruning) by deriving a global importance score LAMP~\cite{lee2020layer} from unstructured pruning to structured pruning. In SP-LAMP, each layer includes a filter with an SP-LAMP score of 1, and the remaining filters are grouped. We utilize a group knapsack solver to maximize the SP-LAMP score under latency constraints. In addition, we improve the strategy of collect the latency to make it more accurate. In particular, for ResNet50/ResNet18 on ImageNet and CIFAR10, SP-LAMP is $1.28\times$/$8.45\times$ faster with $+1.7\%$/$-1.57\% $ top-1 accuracy changed, respectively. Experimental results in ResNet56 on CIFAR10 demonstrate that our algorithm achieves lower latency compared to alternative approaches while ensuring accuracy and FLOPs.

\end{abstract}

\section{Introduction}
Deep Neural Networks (DNNs)~\cite{Ba2013DoDN,Denton2014ExploitingLS,Simonyan2014VeryDC} have achieved remarkable success in recent years, delivering state-of-the-art performance on various tasks, including classification~\cite{Krizhevsky2012ImageNetCW,Guo2018DoubleFP,Srivastava2015TrainingVD}, detection~\cite{Girshick2013RichFH,Redmon2015YouOL,Mao2018TowardsRO,Carion2020EndtoEndOD}, and segmentation~\cite{Shelhamer2014FullyCN,Chen2016DeepLabSI,Chen2018EncoderDecoderWA}. However, their large model sizes and high computational requirements make them challenging to deploy on practical computing systems, particularly on embedded system and Internet of Things devices. To generate more efficient models, researchers have explored various techniques for compressing DNNs without sacrificing their accuracy, such as pruning~\cite{han2015learning,Li2016PruningFF,Yu2019AutoSlimTO,Evci2019RiggingTL,Blalock2020WhatIT,luo2020autopruner}, quantization~\cite{han2015deep,Cai2020ZeroQAN,Wang2020APQJS,Esser2019LearnedSS}, knowledge distillation~\cite{Hinton2015DistillingTK,zhangideal,zhang2022dense}, and matrix low-rank approximation~\cite{zhang2015efficient,Yang2017TensorTrainRN,Yin2020CompressingRN,Yin2021TowardsET}. Among these approaches for model compression, network pruning is a highly effective method that has achieved state-of-the-art performance. 

Previous research on pruning \cite{Li2016PruningFF,Yu2019AutoSlimTO} has mainly focused on reducing the number of parameters and FLOPs~\footnote{The definition of FLOPs is from~\cite{Zhang2017ShuffleNetAE}, i.e., Floating Point Operations.}in DNNs. However, Tan~\etal~\cite{tan2019mnasnet} have revealed that the latency of DNNs can vary greatly, even for models with similar FLOPs. Recently, some reseachers~\cite{chen2018constraint, yang2018netadapt} begin to focus on reducing latency. However, these hardware-friendly pruning methods overlook the latency-accuracy trade-off and hardware latency patterns.

To address the issue in latency-aware pruning, HALP~\cite{shen2021halp} (Hardware-Aware Latency Pruning)introduce a knapsack algorithm to balance accuracy and latency, however it has three limitations. Firstly, it uses the gradient of batch normalization layers as a global importance evaluation metric, which can not provide global filters' information. Secondly, we proof that when HALP collects latency of each filter removed, it does not consider the residual connection in the network and the influence of the position of each layer in the network. Thirdly, HALP use 0-1 knapsack algorithm to solve the latency pruning problem can lead to the removal of all filters in an entire layer.

In this paper, we propose a global importance score SP-LAMP (Structured Pruning-Layer-Adaptive Magnitude-based Pruning). We extend LAMP (Layer-Adaptive Magnitude-based Pruning) from unstructured pruning to structured pruning, which taking into account the connection of each two layer. We employ a group knapsack solver that incorporates latency as a constraint, avoiding removing all filters in the HALP pruning method. This approach offers a more flexible and adaptable pruning technique, allowing for resource allocation while considering the latency constraint of the hardware.

Our main contributions are summarized as follows:
\begin{itemize}[leftmargin=1cm]
\item We propose SP-LAMP method to extend the global importance score LAMP from unstructured pruning to structured pruning, which considering the connection of each layer.
\item We improve latency collection strategy that takes into account the skip connection and filters in different position.
\item We use the grouping knapsack algorithm to replace the 0-1 knapsack algorithm in HALP to conduct structured pruning that preserves the entire layer and produces more flexible pruning results. 
\end{itemize}

\section{Related Work } 
\subsection{Prunning Method}
Pruning methods are widely used in neural networks to solve the performance degradation caused by over-parameterization. Weight pruning focuses on properly selecting the to-be-pruned weights within the filters. OBD and OBS~\cite{lecun1989optimal,hassibi1992second} are classical loss-guided methods, reducing the number of connections based on the second derivative of the loss function. Molchanov~\etal~\cite{Molchanov2019ImportanceEF}  and Han~\etal~\cite{han2015learning}  propose pruning algorithms based on neuron importance ranking. LAMP~\cite{lee2020layer} proposes a novel importance score based on a model-level distortion minimization perspective.

Although enabling a high compression ratio, this strategy meanwhile causes unstructured sparsity patterns, which are not well supported by the general-purpose hardware in practice. So structured pruning ~\cite{wen2016learning,luo2017thinet,min20182pfpce,he2017channel,zhang2018adam,Han2015LearningBW,Liu2017LearningEC,Sui2021CHIPCI,Hou2022CHEXCE} have become a popular stream. 
\subsection{Latency-aware Compression}
Several Neural Architecture Search (NAS) techniques ~\cite{dai2019chamnet,dong2018dpp,tan2019mnasnet,wu2019fbnet} have incorporated platform constraints to reduce latency, but they come at a high computational cost. Other methods, such as constrained Bayesian optimization ~\cite{chen2018constraint} and NetAdapt ~\cite{yang2018netadapt}, have made strides in optimizing for latency, but they do not consider hardware features. HALP~\cite{shen2021halp} uses the 0-1 Knapsack algorithm to solve the latency-constrained structured pruning problem, but the resultant solution can entail removing all neurons in a layer. It is important to note that HALP considers only the latency of each neuron in the current layer, without accounting for skip connections or the potential impact on other layers.

\section{Method}

\begin{figure}[t]
\centering
\includegraphics[width=\textwidth]{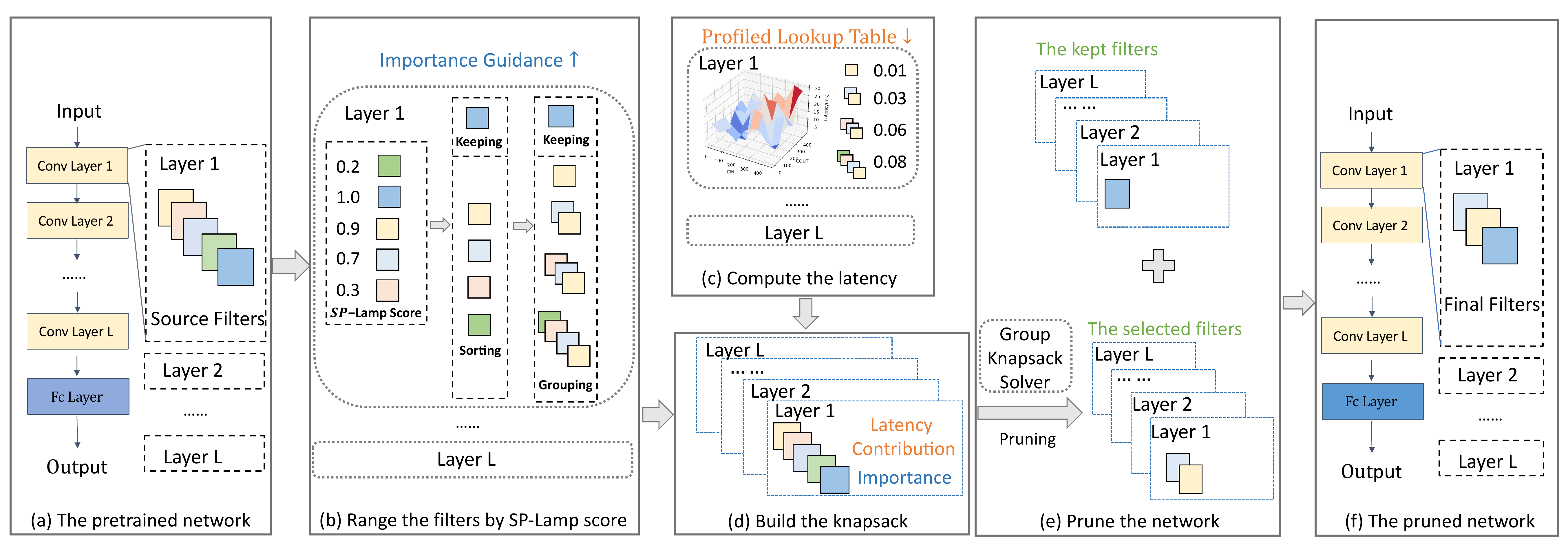}
\caption{An illustration of pruning process of SP-LAMP.(a) The pretrained network and the source filters of each layer; (b) The source filters are sorted based on their SP-LAMP scores, and only the filters with a score of 1 are retained. The remaining filters are then grouped together; (c) Compute the latency contribution of each group based on the latency lookup table; (d) Build knapsack for the filter groups; (e) Utilize the group knapsack solver to determine the optimal selection of filters to retain for each layer. And The filters selected through solving the filters with an SP-LAMP score of 1 in each layer and the knapsack algorithm are considered as the final set of retained filters; (f) The pruned network obtained by SP-LAMP method.}
\label{process}
\vspace{-2mm}
\end{figure}

We depict the workflow of our proposed efficient layer-adaptive structured pruning method guided by latency in Figure~\ref{process}. Our approach incorporates two crucial metrics: the SP-LAMP score and the latency reduction contribution. To implement this methodology, we leverage a group knapsack solver to determine the filters that should be retained before the pruning process. This solver enables us to achieve an optimized selection of filters, ensuring the preservation of essential filters while effectively reducing latency.
\subsection{Generalize LAMP to SP-LAMP}

LAMP~\cite{lee2020layer} proposes a novel importance score based on a model-level distortion minimization perspective. Assuming weights are sorted in ascending order according to the index map, LAMP apply this to each unrolled vector without loss of generality, i.e., $\left | W\left [ u \right ]  \right | \le \left | W\left [ v \right ]  \right |$ holds whenever $u \le v$,
where $\left | W\left [ u \right ]  \right |$ denote the entry of W mapped by the index $u$.
The LAMP score for the $u$-th index of the weight tensor $W$ is then defined as:
\begin{equation}
\label{lamp}
\operatorname{score}(u ; W):=\frac{(W[u])^{2}}{\sum_{v \geq u}(W[v])^{2}}.
\end{equation}
Next, we will generalize LAMP from unstructured pruning to structured pruning by formula derivation.
Given a training set of $n$ instances, $\left\{\left(\mathbf{x}_{j}, \mathbf{y}_{j}\right)\right\}_{j=1}^{n}$, and a well-trained deep neural network of $L$ layers. Denote the input and the output of the whole deep neural network by $\mathbf{X}=\left[\mathbf{x}_{1}, \ldots, \mathbf{x}_{n}\right] \in \mathbb{R}^{d \times n}$ and $\mathbf{Y} \in \mathbb{R}^{n \times 1}$, respectively.

For a layer $l$, we denote the input and output of the layer by $\mathbf{Y}^{l-1}=\left[\mathbf{y}_{1}^{l-1}, \ldots, \mathbf{y}_{n}^{l-1}\right] \in \mathbb{R}^{m_{l-1} \times n}$~\footnote{For simplicity in presentation, we suppose the neural network is a fully-connected network. For convolutional neural networks, $\mathbf{Y}^{l-1}=\left[\mathbf{y}_{1}^{l-1}, \ldots, \mathbf{y}_{n}^{l-1}\right] \in \mathbb{R}^{m_{l-1} \times n \times h_{l-1} \times w_{l-1}}$ which $h_{l-1}$ and $w_{l-1}$ denotes the height and width of the $l-1$ th layer's feature map.} and $\mathbf{Y}^{l}=\left[\mathbf{y}_{1}^{l}, \ldots, \mathbf{y}_{n}^{l}\right] \in \mathbb{R}^{m_{l} \times n}$, respectively, where $\mathbf{y}_{i}^{l}$ can be considered as a representation of $\mathbf{x}_{i}$ in layer $l$, and $\mathbf{Y}^{0}=\mathbf{X}, \mathbf{Y}^{L}=\mathbf{Y}$, and $m_{0}=d$. Using one forward-pass step, we have $\mathbf{Y}^{l}=\sigma\left(\mathbf{Z}^{l}\right)$, where $\mathbf{Z}^{l}=\mathbf{W}_{l}^{\top} \mathbf{Y}^{l-1}$ with $\mathbf{W}_{l} \in \mathbb{R}^{m_{l-1} \times m_{l}}$ \footnote{Similarly, for convolutional neural networks, $\mathbf{W}_{l} \in \mathbb{R}^{m_{l-1} \times m_{l} \times k_{l} \times k_{l}}$, which $k_{l} \times k_{l}$ denotes the shape of kernel.} being the matrix of parameters for layer $l$, and $\sigma(\cdot)$ is the activation function. For convenience in presentation and proof, we define the activation function $\sigma(\cdot)$ as the rectified linear unit (ReLU).

Then we denote $W_{l,k}$ as the matrix of the k-th filter in l-th layer. $\widehat{W}_{l,k}$ as the matrix of the k-th channel in l-th layer.
We sort the filters of the l-th layer in an ascending order according to $\|W_{l,k}\|^{2}_{F} \times \|\widehat{W}_{l+1,k}\|^{2}_{F}$, i.e., $\|W_{l,u}\|^{2}_{F} \times \|\widehat{W}_{l+1,u}\|^{2}_{F} \leq \|W_{l,v}\|^{2}_{F} \times \|\widehat{W}_{l+1,v}\|^{2}_{F}$ holds whenever $u < v$, where $\|\cdot\|_{F}$ is the Frobenius Norm. 

We denote the LAMP score of structured pruning(SP-LAMP) for the u-th index of the layer $l$ ($1 \leq l \leq L-1$) as:

\begin{equation}
\label{sp-lamp}
\operatorname{score}(u,l):=\frac{\|W_{l,u}\|^{2}_{F} \times \|\widehat{W}_{l+1,u}\|^{2}_{F}}{\sum_{v \geq u}{\|W_{l,v}\|^{2}_{F} \times \|\widehat{W}_{l+1,v}\|^{2}_{F}}}.
\end{equation}

For the last layer, the SP-LAMP score is defined as:
\begin{equation}
\label{sp-lamp-L}
\operatorname{score}(u,L):=\frac{\|W_{L,u}\|^{2}_{F}}{\sum_{v \geq u}{\|W_{L,v}\|^{2}_{F}}}.
\end{equation}

The SP-LAMP score measures the importance of a filter among the surviving filters in the same layer, and is defined by Equation~\ref{sp-lamp} and ~\ref{sp-lamp-L}. We use the SP-LAMP score to globally prune filters until the desired sparsity constraint is met, which is equivalent to performing MP with an automatically selected layerwise sparsity.

It is important to note that each layer retains at least one filter with an SP-LAMP score of 1, ensuring survival. SP-LAMP can be extended to structured pruning by considering subsequent layer weights. It is computationally efficient, implemented with basic tensor operations, and requires no hyperparameter tuning.
For more details, see Appendix A.

\subsection{Knapsack Solver \label{sec:nn}}
In this section, we formulate the pruning process as an optimization process. Firstly, we elaborate objective function of pruning optimization process. Then, we will describe the method of collecting latency lookup table and the calculation method of latency contribution. Finally, we elaborate on the process of using the grouping knapsack solver to solve the optimization problem after grouping the filters of each layer.

\noindent\textbf{Objective Function} Consider the problem of network pruning with a given constraint of latency $C$, the optimization problem can be generally formulated as:

\begin{equation}
\underset{\widetilde{W}_{1:L}}{\arg \min } \mathcal{L}(\widetilde{W}_{1:L}, X, Y) \quad \text { s.t. } \quad \Phi\left(f\left(x_{i}; \widetilde{W}_{1:L}\right)\right) \leq C,
\end{equation}
where $\mathcal{L}$ is the loss of the task and $\Phi(.)$ maps the network to the constraint of latency $C$. To address this issue, it is critical to identify the network segment that satisfies the constraint while minimizing performance disruption. Our research demonstrates the effectiveness of the SP-LAMP score as an evaluation metric for quantifying the impact of filter removal on the overall network.
\begin{equation}
\label{optimization}
\begin{aligned}
&\underset{p_{1}, \cdots, p_{l}}{\arg \max } \sum_{l=1}^{L} \sum_{i=1}^{p_{l}} score(p_{l,i},l), 
&\text { s.t. } \quad \sum_{l=1}^{L} T_{l}\left(p_{l}\right) \leq C, \forall l \quad 0 \leq p_{l} \leq m_{l},
\end{aligned}
\end{equation}
where $p_{l}$ denotes the number of kept neurons at layer $l$, and $T_{l}\left(p_{l}\right)$ checks on the associated latency contribution of layer $l$ with $p_{l}$ output channels. $\mid p_{0}$ denotes a fixed input channel number for the first convolutional block, e.g., 3 for RGB images. We next elaborate on $T(\cdot)$ in detail.

\noindent\textbf{Latency Contribution} We empirically obtain the layer latency $T_{l}\left(p_{l}\right)$ in Equation~\ref{optimization} by pre-building a layer-wise look-up table with pre-measured latencies. This layer latency corresponds to the aggregation of the neuron latency contribution of each neuron in the layer, $c_{l}^{j}$ :
\begin{equation}
T_{l}\left(p_{l}\right)=\sum_{j=1}^{p_{l}} c_{l}^{j}, \quad 0 \leq p_{l} \leq m_{l}.
\end{equation}
The latency contribution of the $j$-th neuron in the l-th layer can also be computed using the entries in the look up table as:
\begin{equation}
\label{get-c}
\begin{aligned}
    &c_{l}^{j}= \Phi\left(f\left(x ; W_{(1: l-1)}, \widetilde{W}_{l,\sum j},\overline{W}_{l+1,\sum j}, W_{(l+2: L)}\right)\right) \\
    &- \Phi\left(f\left(x ; W_{(1: l-1)}, \widetilde{W}_{l,\sum j-1},\overline{W}_{l+1,\sum j-1}, W_{(l+2: L)}\right)\right) 
    & 1 \leq j \leq p_{l}.
\end{aligned}
\end{equation}
\noindent\textbf{Latency Lookup Table} It should be notice that our computation of each filter's latency contribution is different from HALP~\cite{shen2021halp}. They only consider the decrease of latency of filter pruning on the current layer and the next layer: $c_{l}^{j}=T_{l}\left(p_{l-1}, j\right)-T_{l}\left(p_{l-1}, j-1\right), \quad 1 \leq j \leq p_{l}$. However, this scheme has two disadvantages: (1) Skip connections are not considered. The removed filters will influence all the layers connected to it. (2) Their calculation method assumes that the latency of each layer is independent, and pruning a certain layer will not affect the layers unrelated to it. However, the actual latency can vary due to differences in hardware and inference engines. For example, removing the same number of filters from layers with the same structure but located at different positions in the network can still result in different latencies.
\begin{figure}[t]
	\centering
		\centering
		\includegraphics[width=0.32\textwidth]{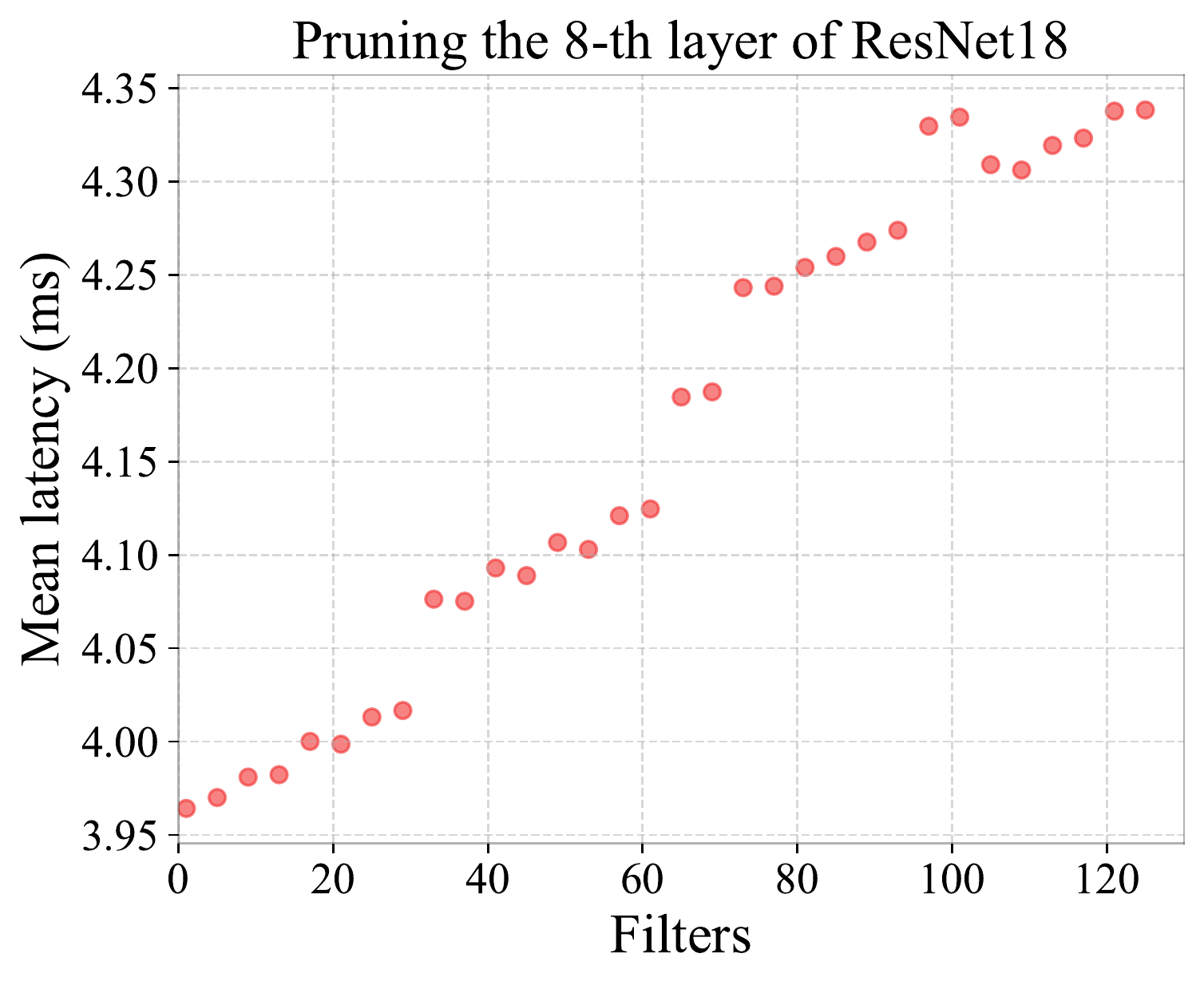}
        \includegraphics[width=0.32\textwidth]{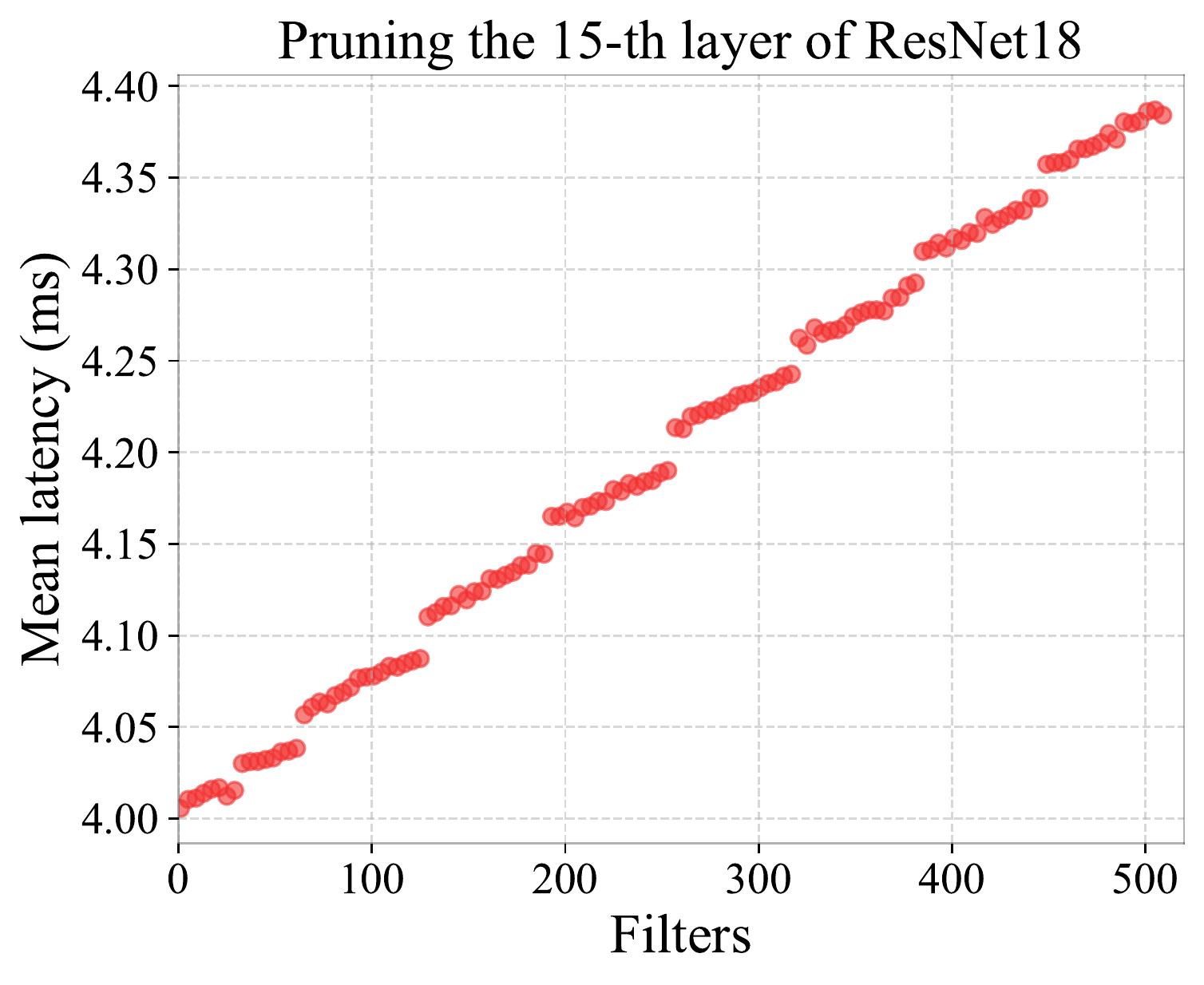}
        \includegraphics[width=0.32\textwidth]{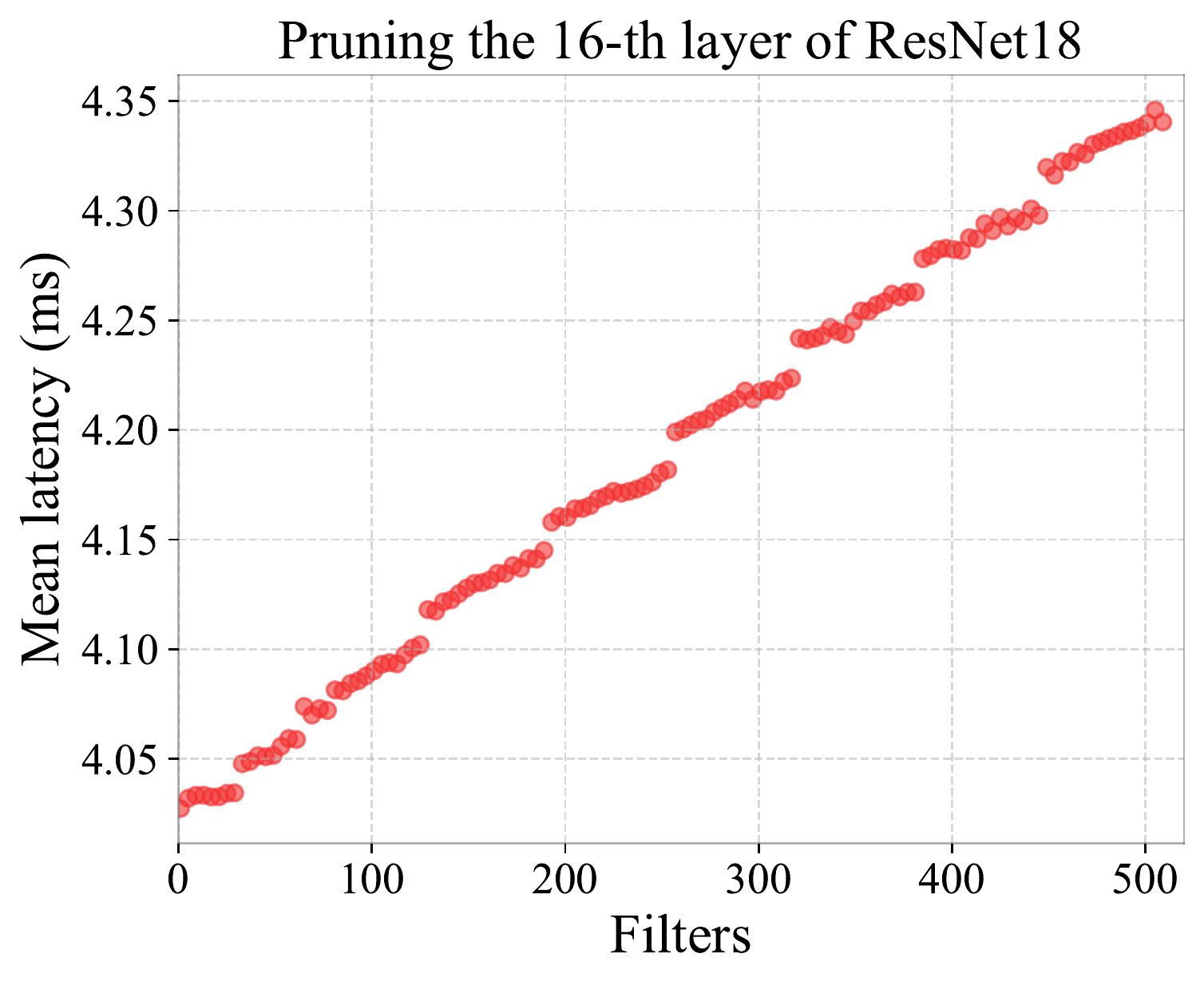}
	\caption{The trend curve of the 8,15,16-th layer of ResNet18 inferred on 2080Ti, every four filters removed corresponds to
one point in the image. We can observe that the decrease of latency brought about by the removal of filters will cause a ”jump”
after a certain number of times.}
	\label{latency_contribution}
 \vspace{-5mm}  
\end{figure}
In order to obtain a relatively accurate $c_{l}^{j}$ in Eq.\ref{get-c} through the following steps: 
\begin{itemize}[leftmargin=1cm]
\item \textbf{Step 1}. Filters are removed individually from each layer, resulting in passive pruning of all affected layers. This process generates a new network structure, which is subsequently inferred by the equipment.
\item \textbf{Step 2}. During inference, all other processes on the equipment are halted, and the utilization rate is maximized to 100\% by adjusting the batch size value.
\item \textbf{Step 3}. We will preheat the equipment, for each sub-model after pruning, we will infer it for 3000 times, and only record the mean and standard deviation of 2500 inferences in the back.
\end{itemize}
We also simplify the process using nn-Meter\cite{Zhang2021nnMeterTA}, which accurately predicts the model's latency and quickly get the predicted latency. In Fig.\ref{latency_contribution}, we show the trend curve of the 8,15,16-th layer of ResNet18 inferred on 2080Ti. The trend curve of the 24th, 44th, and 55th layers of ResNet56 predicted by nn-Meter is included in the Appendix B.
It is observed that pruning a filter can sometimes lead to increased latency in model inference, which may be due to system interference. As a result, some latency contribution values ($c_{l}^{j}$) may be negative.

\noindent\textbf{Group Knapsack Solver} Given both importance and latency estimates, we now aim at solving Equation~\ref{optimization}. By plugging back in the layer importance SP-LAMP (Equation~\ref{sp-lamp}) and layer latency (Equation ~\ref{get-c}), we come to:

\begin{equation}
\label{final}
\begin{aligned}
&\underset{p_{1}, \cdots, p_{l}}{\arg \max } \sum_{l=1}^{L} \sum_{i=1}^{p_{l}} score(p_{l,i},l), 
&\text { s.t. } \quad \sum_{l=1}^{L} \sum_{j=1}^{p_{l}} c_{l}^{j} \leq C,  \quad 0 \leq p_{l} \leq m_{l}.
\end{aligned}
\end{equation}

The SP-LAMP scores of the filters in each layer are sorted in descending order, which simplifies the pruning process into a knapsack problem with additional constraints. In HALP\cite{shen2021halp}, the author regards each filter as an independent object, the solution of 0-1 knapsack problem is adopted. However, this scheme may delete all the filters of the one layer. Therefore, they need to add post-processing to judge whether pruning is legal. To address this problem, we use the grouping knapsack algorithm instead. We treat each layer as a group, for each group, there exists exactly one filter with the SP-LAMP score of 1, which is the maximum SP-LAMP score possible. This filter must be retained so we do not bring it into the calculation of grouping knapsack algorithm. After we arranged the filters of layer $l$ in the order from large to small. Then we group the subsets of the filters together in layer $l$ as an independent item in the knapsack by the following rules:
\begin{equation}
    S_{l,1} = \left\{W_{l,2}\right\} \quad S_{l,i} = S_{l, i-1} \cup \left\{W_{l,i+1}\right\}.
\end{equation}

\begin{algorithm}
\KwData{input data: Latency constaint $C$. Layers of deep neural networks $L$. $I_{l,i}$ denotes the SP-LAMP score of $i$-th group of layer $l$ . Total group num $N_l$ of layer $l$. $S_{l,i} $ is the $i$-th group set of layer $l$. The filters of each layer which SP-LAMP score is equal to 1 is kept in $keeping[l]$. $G_{l,i}$ is a set of filters in $i$-th group in layer $l$. $f[l][d]$ is a two-dimensional array used to record intermediate results}
\KwResult{output data: Kept important filters $select$.}
Initialization\: $f[0,k]\leftarrow 0$\;$select\leftarrow \emptyset $\;
    \For{$l \leftarrow 1$ \KwTo $L+1$}{
  \For{$c \leftarrow 1$ \KwTo $C+1$}{
   $f[l,c] \leftarrow f[l-1,c]$\;
   
    \For{$i \leftarrow 1$ \KwTo $N_l$}{
    \If {$c>=C_{l,i}$}{$f[l,c] = \max(f[l-1,c],f[l-1,c-C_{l,i}]+I_{l,i})$\;}
    }
    }
    }
    \For{$l \leftarrow L$ \KwTo $1$}{ 
    \If{$f[l,C]>f[l-1,C]$}{
   \For{$i \leftarrow 1$ \KwTo $N_l$}{
   $select[l] \leftarrow G_{l,i}$\;
   
   $C \leftarrow C-C_{l,i}$\;
  }
 }}
    \For{$l \leftarrow 1$ \KwTo $L$}{
$select[l] \leftarrow select[l]+ keeping[l]$\;
 }
\caption{Group Knapsack Solver}
 \label{groupk}
\end{algorithm}

Then we can compute the sum of latency $C_{S_{l,i}}$ and SP-LAMP score $I_{S_{l,i}}$ of each item by adding the latency and SP-LAMP of each filter in the subset together separately. Group knapsack is a dynamic programming problem, here, we utilize the following dynamic transfer equation to solve this problem:
\begin{equation}
\begin{aligned}
    &f[l][v]=Max\left\{\begin{array}{l}
    f[l-1][v] \\
    f[l-1][v-C_{S_{l,i}}]+I_{S_{l,i}}
\end{array}\right. 
    &\text { s.t. } \quad 0 \leq v \leq C,  \quad 1 \leq l \leq L,
\end{aligned}
\end{equation}
where $f[l-1][v]$ is a two-dimensional array used to record intermediate results. We can obtain $f[L][C]$ as the final result and obtain the final pruned network by backward deduction. 

A detailed description of the pseudo code of the group knapsack solver is provided in Algorithm~\ref{groupk}.
\subsection{Layer-adaptive Structured Pruning Guided by Latency}
We finally implement our proposed pruning method as follows:
Our pruning process takes the pretrained network as input and uses latency $C$ as the constraint to obtain a pruned network. To simplify the implementation of the knapsack algorithm, we will scale the value of $C$ by multiplying it by 10000. This adjustment is made because $C$ is typically measured in milliseconds, which can be challenging to handle directly in the calculations. In particular, we set up k stages to gradually reduce latency to meet latency requirements, with $C_1$,...,$C_k$, $C_k=C/k$. The algorithm gradually prune filters using steps below:
\begin{itemize}[leftmargin=1cm]
\item \textbf{Step 1.} We firstly rank filters in each layer by decending SP-LAMP score as Equation~\ref{sp-lamp}. Notably, filters with a SP-LAMP score of 1 are retained. Then we group the remaining filters for each layer. The groups of each layer can vary from having only one filter to including all filters.
\item \textbf{Step 2.} We count the number of filters remaining in each group of layers and get the latency contribution of each group as in Equation~\ref{get-c}.
\item \textbf{Step 3.} We execute the Algorithm~\ref{groupk} to select the filters being remained with current latency constraint $C_i$. Repeat starting from the Step 1 until latency constraint C are reached. 
\end{itemize}
We keep the selected filters and filters with SP-LAMP score equal to 1 as the final selected filters in pruned network. Once pruning finishes we fine-tune the network to recover accuracy.
\begin{table}[htpb]

    \caption{Pruning results of ResNet-50 and MobileNet on ImageNet.}
    \centering
\resizebox{\linewidth}{!}
{
\scalebox{0.6}{
\begin{tabular}{@{}lllllllll@{}}
\toprule
\multicolumn{1}{c}{\multirow{2}{*}{Model}} & \multicolumn{1}{c}{\multirow{2}{*}{Method}} & Base  & Base  & Pruned & Pruned & \multirow{2}{*}{Top-1 $\downarrow$ } & \multirow{2}{*}{Top-5 $\downarrow$ } & \multirow{2}{*}{FLOPs $\downarrow \%$} \\
\multicolumn{1}{c}{}                       & \multicolumn{1}{c}{}                        & Top-1 & Top-5 & Top-1  & Top-5  &                        &                        &                           \\ \midrule
\multirow{24}{*}{ResNet-50}                & SFP~\cite{he2018soft}                                        & 76.15      &  92.87     & 74.61       & 92.06       &  1.54                      &  0.81                      &   41.80                        \\
                                           & GAL-0.5~\cite{lin2019towards}                                    & 76.15      & 92.87      &  71.95      &  90.94      &   4.20                     &    1.93                    &  43.03                         \\
                                           & NISP~\cite{yu2018nisp}                                       &   -    &  -     & -       &  -      & 0.89                       &  -                      &    44.01                       \\
                                           & Taylor-FO-BN~\cite{molchanov2019importance}                               &  76.18     &  -     &  74.50      & -       &   1.68                     & -                      &  44.98                         \\
                                           & Channel Pruning~\cite{he2017channel}                                 & -      & 92.20      &  -      &  90.8      &  -                      &    1.40                    &   50.00                        \\
                                           & HP~\cite{xu2018hybrid}                                        & 76.01      & 92.93      & 74.87       & 92.43       &  1.14                      &  0.50                      &  50.00                         \\
                                           & MetaPruning~\cite{liu2019metapruning}                                & 76.6      &  -     &  75.4      & -       & 1.20                       &  -                      &  51.50                         \\
                                           & Autopr~\cite{luo2020autopruner}                                      & 76.15      & 92.87      &  74.76      &  92.15      &   1.39                     &  0.72                      &     51.21                      \\
                                           & GDP~\cite{lin2018accelerating}                                        & 75.13      &  92.30     &  71.89      & 90.71       &    3.24                    & 1.59                       &  51.30                         \\
                                           & FPGM~\cite{he2019filter}                                        & 76.15      & 92.87      &   74.83     &  92.32      &   1.32                     &  0.55                      & 53.50                          \\
                                           & CHEX~\cite{Hou2022CHEXCE}                                     &  77.80   &-  & 77.40           &   -                        &   0.04       &-              &   51.20                        \\
                                            & RL-MCTS~\cite{Wang2022ChannelPV}                                     &  77.34   &93.37 & 76.46           &   92.83    &   0.88                    &   0.54                     &   55.00                       \\
                                           
                                           &\textbf{Ours}-55\%                                        &    \textbf{76.15}    & \textbf{92.87}       &  \textbf{76.12}      &   \textbf{92.85}     &   \textbf{0.03}                     &            \textbf{0.02}            &   \textbf{55.37}                        \\
                                           & C-SGD (extension)~\cite{ding2021manipulating}                          & 76.15      &   92.87    &  75.29      & 92.39       &  0.86                      &   0.48                     &  55.44                         \\
                                           & DCP~\cite{zhuang2018discrimination}                                        &  76.01     &  92.93     &  74.95      &  92.32      &    1.06                    &  0.61                      &    55.76                       \\
                                           & C-SGD~\cite{ding2019centripetal}                                     & 75.33      &  92.56     &  75.54      &  92.09      &   0.79                     &   0.47                     &    55.76                       \\
                                           & ThiNet~\cite{luo2017thinet}                                   & 75.30      & 92.20      & 72.03        &  90.99      &    3.27                    &  1.21                      &   55.83                        \\
                                           & SASL~\cite{shi2020sasl}                                      & 76.15      &  92.87     &  75.15      &  92.47      &   1.00                     &       0.40                 &    56.10                       \\
                                           & TRP ~\cite{xu2020trp}                                      &   75.90    &  92.70     &  72.69      &   91.41     &     3.21                   &    1.29                    &    56.52                       \\
                                           & \textbf{Ours}-60\%                                         &   \textbf{76.15}    &  \textbf{92.87}     &  \textbf{76.04}      &   \textbf{92.58}     &     \textbf{0.11}                   &    \textbf{0.29}                    &    \textbf{60.14}                       \\

                                           & GAL ~\cite{lin2019towards}                                       & 76.15      &  92.87     &  69.31      &   89.12     &    6.84                   &     3.75                  &   72.90                        \\ 
                                           & HRank ~\cite{lin2020hrank}                                      &  76.15     &  92.87     &  69.10      &   89.58     &      7.05                  &    3.29                   &      75.88                     \\
                                           & \textbf{Ours}-75\%                                        &    \textbf{76.15}   &  \textbf{92.87}     &  \textbf{70.04}      &  \textbf{89.78}      &     \textbf{6.11}                   &     \textbf{3.09}                   &     \textbf{75.89}                      \\                                           \midrule
\multirow{3}{*}{MobileNetV2}               & MetaPruning~\cite{liu2019metapruning}                                  &   70.60    &  -     &  66.10      &  -      &     4.50                   &    -                    &  73.81                         \\
                                           & ResRep~\cite{ding2021resrep}                                       &  70.78     &  89.78     &   68.02     &   87.66     &     2.76                   &     2.12                   &     73.91                      \\

                                           & \textbf{Ours}-75\%                                        &   \textbf{70.34}    &  \textbf{89.72}     &   \textbf{68.11}     &    \textbf{87.75}    &        \textbf{2.23}                &         \textbf{1.97}               &   \textbf{74.03}                        \\ \bottomrule
\end{tabular}}
}
\label{ImageNet}

\end{table}
\section{Experiment}
\subsection{Experimental Setup and Baselines}
\noindent\textbf{Baselines Models and Datasets.} To demonstrate the effectiveness and generality of our proposed
SP-LAMP approach, we evaluate its pruning performance for various baseline
models on different image classification datasets. To be specific, we conduct experiments for ResNet-50~\cite{he2016deep} and MobileNetV2~\cite{howard2017mobilenets}  on large-scale ImageNet1K~\cite{Deng2009ImageNetAL} dataset . Also, we further
evaluate our approach and compare latency performance with other state-of-the-art pruning methods for ResNet18 and ResNet56 model on  CIFAR10~\cite{krizhevsky2009learning} dataset.

\noindent\textbf{Pruning and Fine-tuning Configurations.}
We follow the data augmentation of PyTorch official example including random cropping and flipping.  For ResNet-50, we use the official torchvision base model (76.15\% top-1 accuracy) for the fair comparison with most competitors. For MobileNet, we train from scratch with an initial learning rate of 0.1, batch size of 512 and cosine learning rate annealing for 70 epochs. The top-1 accuracy is 70.34\%, slightly higher than that reported in the original paper. We train the base models with batch size of 64 and the common learning rate schedule which is initialized as 0.1, multiplied by 0.1 at epoch 120 and 180, and terminated after 240 epochs. We perform the fine-tuning for 200 epochs on CIFAR-10 datasets with the batch size,
 and initial learning rate as 128 and 0.01, respectively.  We count the FLOPs as multiply-adds, which is 4.09G for ResNet-50 , 569M for MobileNet, 128M for ResNet56.
\subsection{Compare with Other Pruning Algorithms}
Table~\ref{ImageNet} show the superiority of our algorithm. The results of other methods are copied from origin papers. Owing to the pruned model of their methods cannot be obtained, we can not compare their latency. On ResNet-50, our algorithm achieves 0.03\% top-1 accuracy drop, which is the first to realize lossless pruning with such high pruning ratio (55.37\%), to the best of our knowledge. In terms of top-1 accuracy drop, SP-LAMP outperforms SASL by 0.89\%, HRank by 0.94\% and all the other recent competitors by a large margin. For FLOPs on MobileNet, our method outperforms MetaPruning by 0.22\%. 

\subsection{Performance of Latency}

\begin{table}[htpb]
\vspace{-2mm}
  \caption{ImageNet structured pruning results. With detailed comparison to state-of-the-art pruning methods over varying performance metrics on ResNet-50 for ImageNet.}
  \label{sample-table}
  \centering
\scalebox{0.8}{
\begin{tabular}{llllll}
\toprule
Method           & FLOPs(G)      & Top1          & Top5           & FPS           & Speedup        \\ \midrule
No pruning       & 4.1           & 76.2          & 92.87          & 1019          & 1$\times$             \\
ThiNet-70~\cite{luo2017thinet}        & 2.9           & 75.8          & 90.67          & -             & -              \\
AutoSlim ~\cite{Yu2019AutoSlimTO}        & 3.0           & 76.0          & -              & 1215          & 1.14$\times$          \\
GReg-1  ~\cite{Wang2020NeuralPV}          & 2.7           & 76.3          & -              & 1171          & 1.15$\times$          \\
HALP-80\%  ~\cite{shen2021halp}      & 3.1           & 77.2          & 93.47          & 1256          & 1.23$\times$          \\
\textbf{Ours}-80\%   & 3.0 & \textbf{77.9} & \textbf{93.81} & \textbf{1307} & \textbf{1.28$\times$} \\ \midrule
0.75 $\times$ ResNet-50 & 2.3           & 74.8          & -              & 1467          & 1.44$\times$          \\
ThiNet-50 ~\cite{luo2017thinet}       & 2.1           & 74.7          & 90.02          & -             & -              \\
AutoSlim ~\cite{Yu2019AutoSlimTO}    & 2.0           & 75.6          & -              & 1592          & 1.56$\times$          \\
MetaPruning  ~\cite{liu2019metapruning}      & 2.0           & 75.4          & -              & 1604          & 1.58$\times$          \\
GBN  ~\cite{You2019GateDG}           & 2.4           & 76.2          & 92.83          & -             & -              \\
CAIE ~\cite{Wu2020ConstraintAwareIE}         & 2.2           & 75.6          & -              & -             & -              \\
LEGR ~\cite{Chin2019TowardsEM}          & 2.4           & 75.6          & 92.70          & -             & -              \\
GReg-2  ~\cite{Wang2020NeuralPV}          & 1.8           & 75.4          & -              & 1414          & 1.39$\times$          \\
HALP-55\% ~\cite{shen2021halp}       & 2.0           & 76.5          & 93.05          & 1630          & 1.60$\times$          \\
\textbf{Ours}-75\%              & 2.3          & 76.1         & 92.98         & \textbf{1695}          & \textbf{1.66}$\times$          \\ \midrule
0.5 $\times$ ResNet-50  & 1.1           & 72.0          & -              & 2498          & 2.45$\times$          \\
ThiNet-30 ~\cite{luo2017thinet}       & 1.2           & 72.1          & 88.30          & -             & -              \\
AutoSlim ~\cite{Yu2019AutoSlimTO}        & 1.0           & 74.0          & -              & 2390          & 2.45$\times$          \\
MetaPruning  ~\cite{liu2019metapruning}      & 1.0           & 73.4          & -              & 2381          & 2.34$\times$          \\
CAIE ~\cite{Wu2020ConstraintAwareIE}            & 1.3           & 73.9          & -              & -             & -              \\
GReg-2 ~\cite{Wang2020NeuralPV}        & 1.3           & 73.9          & -              & 1514          & 1.49$\times$          \\
HALP-30\% ~\cite{shen2021halp}       & 1.0           & 74.3          & 91.81          & 2755          & 2.70$\times$          \\
\textbf{Ours}-50\%    & \textbf{0.87} & 73.6 & 90.03 & \textbf{2934} & \textbf{2.87$\times$} \\
\bottomrule
\end{tabular}}
\label{compare latency}
\end{table}

\begin{figure*}[htpb]

    \centering
    \begin{subfigure}[t]{0.4\textwidth}
           \centering
           \includegraphics[width=\textwidth]{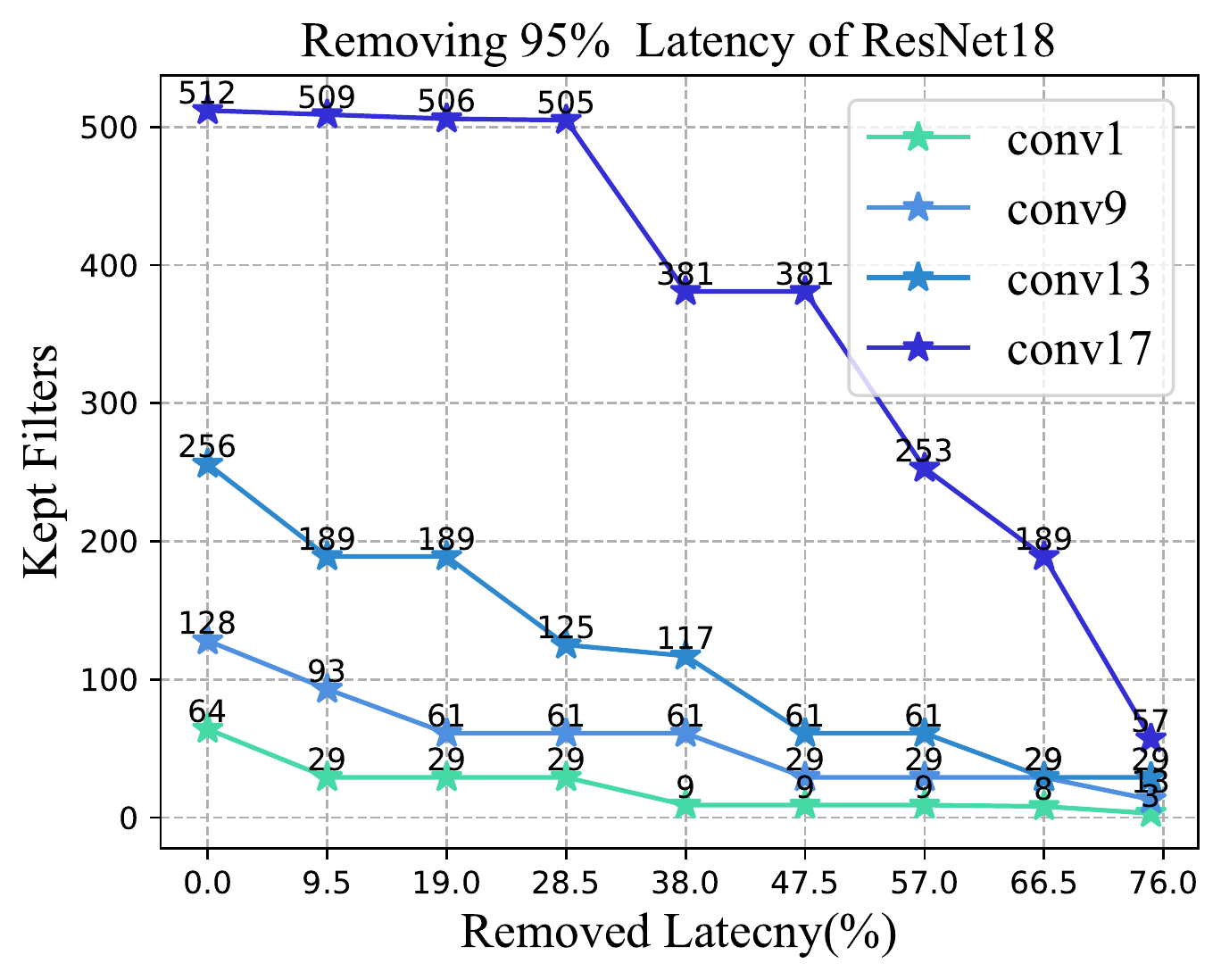}
            \caption{Remove 95\% latency of ResNet18. }
            \label{fig:a}
    \end{subfigure}
    \hspace{.25in}
    \begin{subfigure}[t]{0.4\textwidth}
            \centering
            \includegraphics[width=\textwidth]{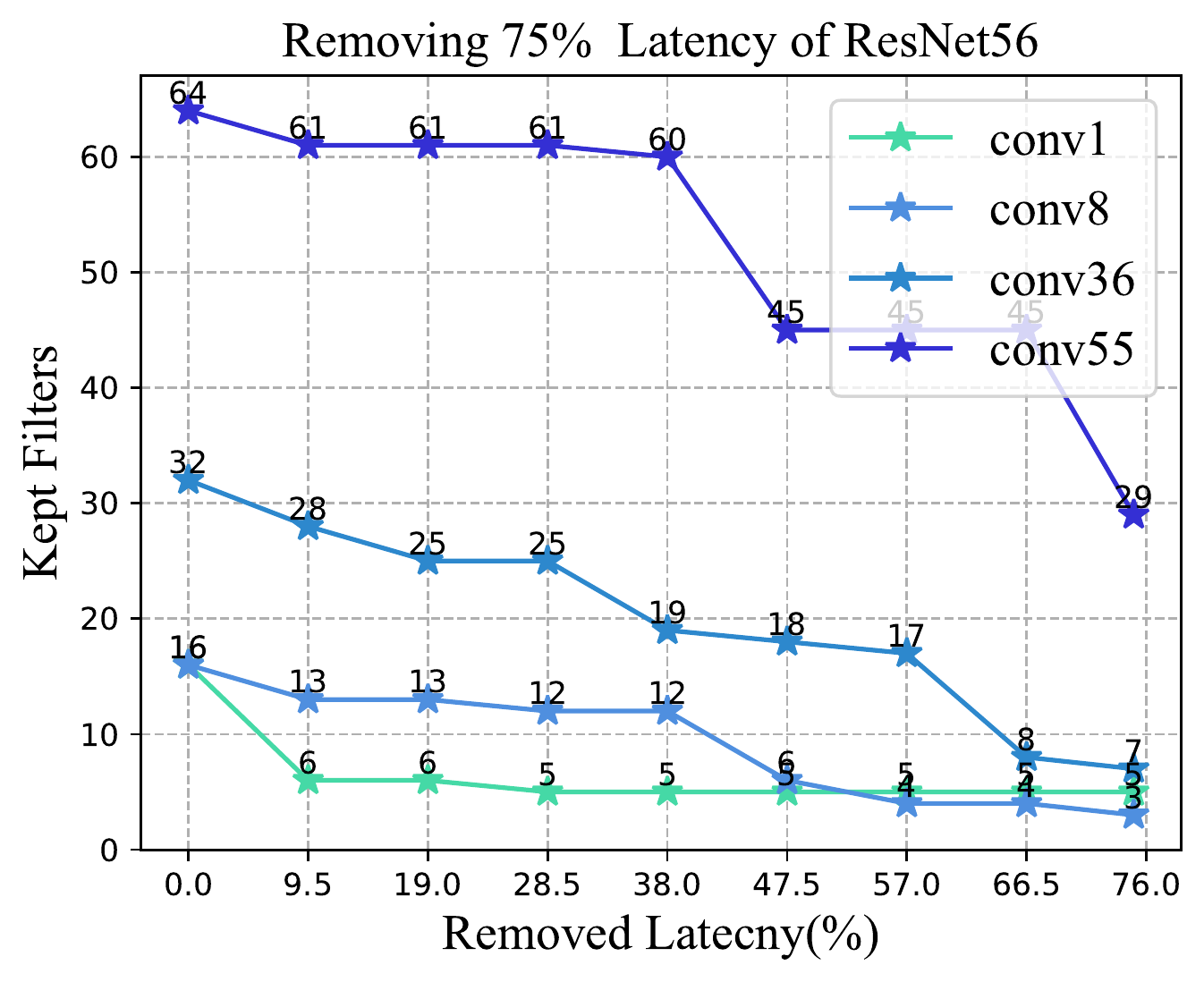}
            \caption{Remove 75\% latency of ResNet56.}
            \label{fig:b}
    \end{subfigure}
    \caption{Our experiments on ResNet18 and ResNet56 show the impact of filter pruning on latency reduction, with 95\% and 75\% of latency removed, respectively. As the proportion of removed latency increases, the number of filters gradually decreases, especially in layers with larger initial filter sizes, indicating significant redundancy.}
\end{figure*}
\noindent\textbf{ResNet50.} We start by pruning ResNet50 and compare our results with state-of-the art
methods in Tab.~\ref{compare latency} on TITAN V. In order to have a fair comparison of the latency, for all the other methods, we recreate pruned networks according to the pruned structures they published and measure the latency. Those methods showing ‘-’ in the table do not have pruned structures published so we are unable to measure the latency. We report FPS (frames per second) in the table and calculate the speedup of a pruned network as the ratio of FPS between pruned and unpruned models. From the results comparison we can find that for pruned networks with similar FLOPs using different methods, our method achieves higher FPS and the fastest inference speed. This also shows that FLOPs do not correspond 1:1 to the latency. For the pruned ResNet50 network with 3G FLOPs remaining, our method achieves a 0.7\% higher top1 accuracy and slightly (0.05$\times$) faster inference.
\begin{table}[t]
    \caption{Pruning results of ResNet-18 on CIFAR-10.}
    \centering
\scalebox{0.8}{
\begin{tabular}{@{}llllllll@{}}
\toprule
\multirow{2}{*}{Method} & Base          & Pruned      & \multirow{2}{*}{Top-1 $\downarrow$ }  &\multirow{2}{*}{Params}    & \multirow{2}{*}{FLOPs}     &  
 \multirow{2}{*}{Speedup }  \\
   & Top-1          & Top-1          &               &        & & &         \\
                         \midrule
Train From Scratch                    & 94.50           & 88.30    &  6.20     & 467K           & 13.36M         &8.45$\times$ \\
ADMM~\cite{Boyd2011DistributedOA}                  & 94.5         & 87.93         & 6.57 &431K	&12.886M 
     &8.5 $\times$     \\
ExpandNets~\cite{Guo2018ExpandNetsLO}                & 94.50          & 89.27          &5.23  &467K	&13.36M  &8.45$\times$
\\
HALP~\cite{shen2021halp}                   & 94.50          & 90.15          & 4.35 &598K	&15.48M 
   &6.46  $\times$    \\

Only Structured Lamp                 & 94.50         & 90.73         & 3.77
& 473K	&17.29M &4.48$\times$
\\

\textbf{Ours}-90\%           & \textbf{94.50} & \textbf{90.96} & \textbf{3.54} & 467K &13.36M&8.45$\times$\\ \bottomrule
\end{tabular}}
\vspace{-4mm}
\label{cifar10}
\end{table}

\begin{figure}[t]
\centering
\includegraphics[width=0.9\textwidth]{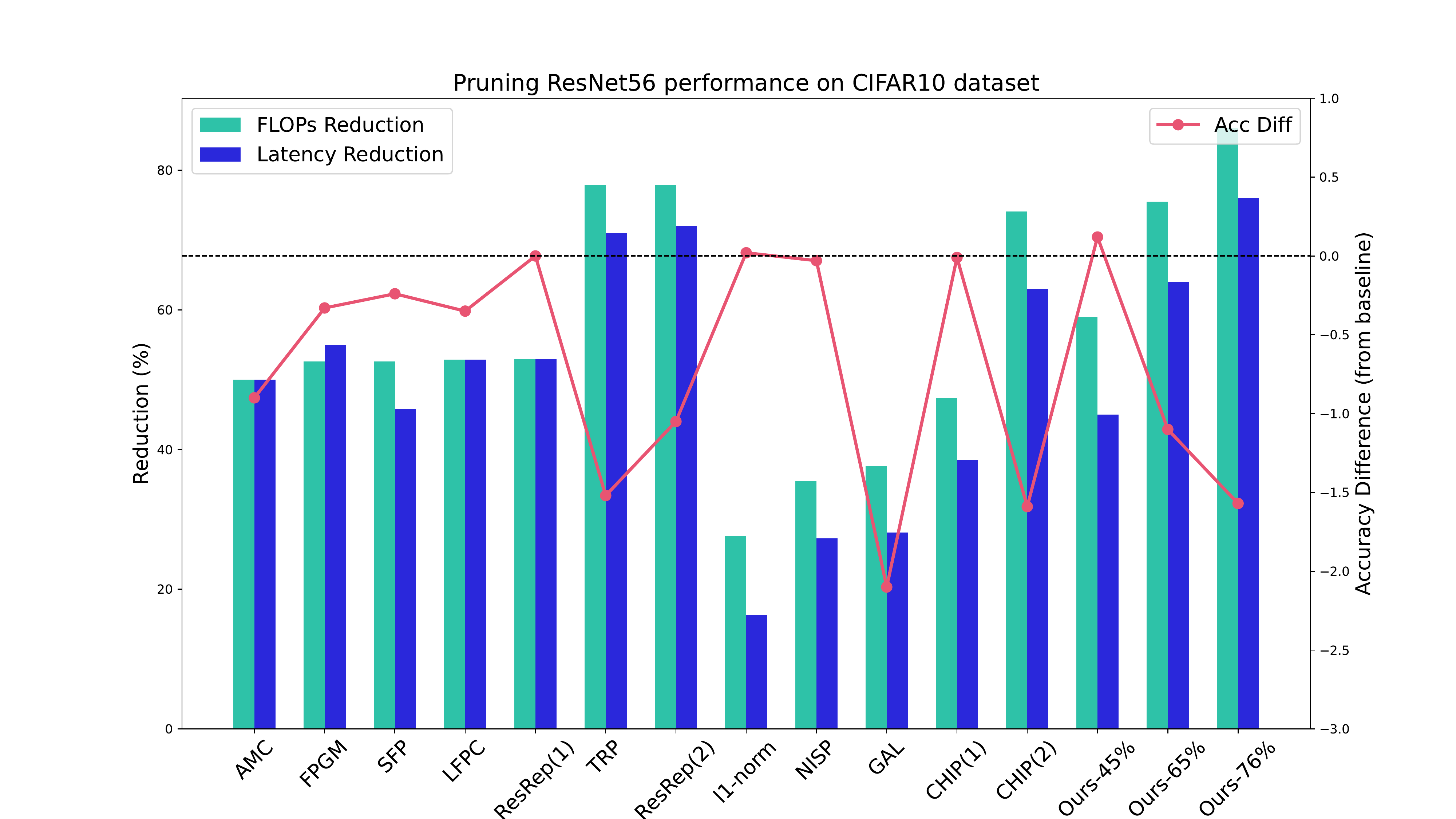}
\caption{We compare the results of pruning ResNet56 with multiple methods ~\cite{he2018amc,he2019filter,he2018soft,he2020learning,ding2021resrep,xu2020trp,Li2016PruningFF,yu2018nisp,lin2019towards,Sui2021CHIPCI} and SP-LAMP with three ratio on the CIFAR10 dataset. The figure depicts the percentage of latency reduction achieved by each method, including our proposed methods FLOPs and nn-Meter, along with the difference in accuracy compared to the base model.}
\label{exp}
\vspace{-7mm}
\end{figure}

\noindent\textbf{ResNet18.} We also report the performance of ResNet18 on the CIFAR10 dataset in Table ~\ref{cifar10}. In order to better illustrate the superiority of our method over others, we report the number of parameters, FLOPs, and  on the 2080Ti. As can be seen, our method achieves the faster speedup despite having only a 3.54\% lower top 1 accuracy than the baseline model. While the ADMM method produces a pruning model with lower latency than our method, our method achieves 3.03\% higher accuracy. The 
\begin{wraptable}{r}{6cm}
\caption{Stage of pruning ResNet56}
\centering
\begin{tabular}{ccc}
\toprule
Stage & Removed Latency \% & ACC     \\
                         \midrule
1 & 9.5\% & 0.00\\
4& 38\% & +2.00\\
5 & 47.5\% & -1.5\\
8 & 85.5\% & -3.5\\
10 & 95.0\% & -9.89\\

\bottomrule
\end{tabular}
\label{remove}

\end{wraptable}
\noindent\textbf{ResNet56.} Figure~\ref{exp} summarizes the pruning performance of our approach for ResNet56 on CIFAR10. During our implementation, we use nn-Meter to construct the latency look up table of ResNet56 and perform latency predictions on pruning models of various methods. From the results comparison we can find that for pruned networks with similar FLOPs using different methods, our method achieves the highest accuracy and also the lower latency. This also shows that FLOPs do not correspond 1:1 to the latency. In Table~\ref{remove}, we remove 95\% latency of ResNet56 on 10 stages and show the corresponding accuracy achieved at each stage.  

We also show the filters change curve of ResNet18 in Figure~\ref{fig:a} and ResNet56 in Figure~\ref{fig:b} gradually removing latency.

\section{Conclusion}
We propose SP-LAMP, a global importance score specifically designed for structured pruning adaptively, taking into account the inter-layer connections. We utilize a group knapsack solver to select filters based on latency constraints. Additionally, we propose a novel strategy for latency collection that considers the position of filters. Experimental results demonstrate the effectiveness of SP-LAMP, showcasing consistent improvements over existing state-of-the-art methods.
\bibliographystyle{unsrt}
\bibliography{main}

\newpage
\appendix
\section{Derivation of SP-LAMP}
During layer-wise pruning in layer l, the input $\mathbf{Y}^{l-1}$ is fixed as the same as the well-trained network. After we remove a filter in layer l (in practice, we set all the values of this filter to 0), we obtain a new output for layer $l$, denoted by $\hat{\mathbf{Y}}^{l}$. Consider the root of mean square error between $\hat{\mathbf{Y}}^{l}$ and $\mathbf{Y}^{l}$ over the whole training data as the layer-wise error: 

\begin{equation}
\varepsilon^{l}=\sqrt{\frac{1}{n} \sum_{j=1}^{n}\left(\left(\hat{\mathbf{y}}_{j}^{l}-\mathbf{y}_{j}^{l}\right)^{\top}\left(\hat{\mathbf{y}}_{j}^{l}-\mathbf{y}_{j}^{l}\right)\right)},
\end{equation}

our goal is to find and remove a filter in layer l to minimize the above equation, so we can rewrite it into:

\begin{equation}
\min \varepsilon^{l} = \arg \min _{q}\sqrt{\frac{1}{n} \sum_{j=1}^{n}\left(\left(\mathbf{y}_{j}^{l,q}\right)^{\top}\left(\mathbf{y}_{j}^{l,q}\right)\right)},
\end{equation}

then we denote $\mathbf{Y}^{l,q} = \left[\mathbf{y}_{1}^{l,q}, \ldots, \mathbf{y}_{n}^{l,q}\right] \in R^{1 \times n}$ as the output feature of the q-th filter in layer l, the above equation equals to:

\begin{equation}
    \arg \min _{q} \sqrt{\frac{1}{n} \sum_{j=1}^{n}\left(\left(\mathbf{y}_{j,q}^{l}\right)^{\top}\left(\mathbf{y}_{j,q}^{l}\right)\right)}=\arg \min _{q} \frac{1}{\sqrt{n}}\left\|\mathbf{Y}^{l,q}\right\|_{F}.
\end{equation}

We have denoted $\mathbf{Z}^{l,q}$ is outcome of the weighted sum operation right before performing the activation function $\sigma(\cdot)$ at the q-th filter in layer $l$ of the well-trained neural network, then we can get:

\begin{equation}
   \arg \min _{q}\frac{1}{\sqrt{n}}\left\|\mathbf{Y}^{l,q}\right\|_{F} \leq \arg \min _{q} \frac{1}{\sqrt{n}}\left\|\mathbf{Z}^{l,q}\right\|_{F},
\end{equation}

then we utlize the Cauchy-Schwarz inequaliy: $\|A x\|_{2}^{2}=\sum_{i}\left(\sum_{j} A_{i j} x_{j}\right)^{2} \leq$ $\sum_{i}\left(\sum_{j} A_{i j}^{2}\right) \cdot\left(\sum_{j} x_{j}^{2}\right)=\|A\|_{F}^{2}\|x\|_{2}^{2}$, where subscripts denote weight indices, we can get:

\begin{equation}
    \begin{aligned}
    \arg \min _{q} \frac{1}{\sqrt{n}}\left\|\mathbf{Z}^{l,q}\right\|_{F} &= \arg \min _{q} \frac{1}{\sqrt{n}}\left\|\mathbf{W}_{l,q} \times x^{l}\right\|_{F} &\leq \arg \min _{q} \frac{1}{\sqrt{n}}\left\|\mathbf{W}_{l,q}\right\|_{F} \left\|x^{l}\right\|_{F}.
    \end{aligned}
\end{equation}

We should notice that, $\frac{1}{\sqrt{n}}$ is a constant and $\left\|x^{l}\right\|_{F}$ is the input of the l-th layer. Then we denote $\left\|\mathbf{W}_{l,q}\right\|_{F}$ as the minimum distortion of the l-th layer after removing a filter.

Building on this speculation,
we now ask the following question: “How can we select the filter to have small
model-level output distortion?” To formalize, we consider the minimization:

\begin{equation}
\label{main_opti}
\arg \min _{q} sup \left\|f\left(x ; W_{(1: L)}\right)-f\left(x ; \widetilde{W}_{(1: L)}\right)\right\|_{F},
\end{equation}

where $\widetilde{W}_{i,q} = \left\|W_{(i)}-W_{(i,q)}\right\|_{F}$ denotes the weight matrix that the q-th filter which has minimization distortion in i-th layer has been removed. $\overline{W}_{i+1,q} = \left\|W_{(i+1)}-\widehat{W}_{(i+1,q)}\right\|_{F}$ denotes the weight matrix that the q-th channel in i+1-th layer has been removed. Due to the nonlinearities from the activation functions, it is difficult to solve Eq. \ref{main_opti} exactly. Similar to LAMP, we consider the following greedy procedure: At each step, we (a) approximate the distortion incurred by pruning a filter, (b) remove the filter with the smallest SP-LAMP score, and then (c) go back to step (a) and re-compute the SP-LAMP scores based on the pruned model.

Once we assume that only one filter is pruned at a single iteration of the step (a), we can use the following upper bound of the model output distortion to relax the optimization (\ref{main_opti}): With $\widetilde{W}_{i}$ denoting a pruned version of $W_{i}$, we have:

\begin{equation}
\label{result}
    \begin{aligned}
    sup & \left\|f\left(x ; W_{(1: L)}\right)-f\left(x ; W_{(1: i-1)}, \widetilde{W}_{i,q},\overline{W}_{i+1,q}, W_{(i+2: L)}\right)\right\|_{F} \\ \leq &\frac{\left\|W_{i,q}\right\|_{F} \left\|\widehat{W}_{i+1,q}\right\|_{F}}{\left\|W_{i}\right\|_{F} \left\|W_{i+1}\right\|_{F} }\left(\prod_{j=1}^{L}\left\|W_{j}\right\|_{F}\right)
    \end{aligned}
\end{equation}

Next, we will prove the above inequality. In order to simplify the following formula derivation, we assume we only prune front l-1 layers(actually, the last layer as the output layer will not be pruned but it can be generalized easily if we prune it). This inequality is a simplified and modified version of what is popularly known as “peeling” procedure in the generalization literature, and we present the proof only for the sake of completeness. We begin by peeling the outermost layer as:

\begin{equation*}
    \begin{aligned}
        &\left\|f\left(x ; W_{(1: L)}\right)-f\left(x ; W_{(1: i-1)}, \widetilde{W}_{i,q}, \overline{W}_{i+1,q}, W_{(i+2: L-1)}\right)\right\|_{F} \\
       =&\left\|W_{L}\left(\sigma\left(f\left(x ; W_{(1: L-1)}\right)\right) \right. \right. \\
        &\left. \left. -\sigma\left(f\left(x ; W_{(1: L-1)}, \widetilde{W}_{i,q}, \overline{W}_{i+1,q}, W_{(i+2: L-1)}\right)\right)\right)\right\|_{F} \\
    \leq& \left\|W_{L}\right\|_{F} \cdot \left\| \sigma\left(f\left(x ; W_{(1: L-1)}\right) \right. \right. \\
        &\left. \left. -\sigma\left(f\left(x ; W_{(1: i-1)}, \widetilde{W}_{i,q}, \overline{W}_{i+1,q}, W_{(i+2: L-1)}\right)\right)\right)\right\|_{F}\\
    \leq&\left\|W_{L}\right\|_{F} \cdot\left\|f\left(x ; W_{(1: L-1)}\right) \right. \\
        &\left. -f\left(x ; W_{(1: i-1)}, \widetilde{W}_{i,q}, \overline{W}_{i+1,q}, W_{(i+2: L-1)}\right)\right\|_{F}  ,
    \end{aligned}
\end{equation*}

where we have used Cauchy-Schwarz inequality for the first inequality, and the 1-Lipschitzness of
ReLU activation with respect to Frobenius Norm. We keep on peeling until we get:

\begin{equation}
\label{peel back}
\begin{aligned}
&\left\|f\left(x ; W_{(1: L)}\right)-f\left(x ; W_{(1: i-1)}, \widetilde{W}_{i,q}, \overline{W}_{i+1,q}, W_{(i+2: L)}\right)\right\|_{F} \\
\leq&\left(\prod_{j>i+1}\left\|W_{j}\right\|_{F}\right) \cdot\left\|f\left(x ; W_{(1: i+1)}\right) \right. \\
& \left. -f\left(x ; W_{(1: i-1)}, \widetilde{W}_{i,q}, \overline{W}_{i+1,q}\right)\right\|_{F}
\end{aligned}
\end{equation}

The second multiplicative term on the right-hand side requires a slightly different treatment. Via
Cauchy-Schwarz, we get:

\begin{equation}
\label{peel i+1}
\begin{aligned}
&\left\|f\left(x ; W_{(1: i+1)}\right)-f\left(x ; W_{(1: i-1)}, \widetilde{W}_{i,q}, \overline{W}_{i+1,q}\right)\right\|_{F}\\
=&\left\|\left(\widehat{W}_{i+1,q}\right) \sigma\left(f\left(x ; W_{(1: i)}\right)-   f\left(x ; W_{(1: i-1)},\widetilde{W}_{i,q}\right)\right)\right\|_{F} \\
\leq&\left\|\widehat{W}_{i+1,q}\right\|_{F} \cdot\left\|\sigma\left(f\left(x ; W_{(1: i)}\right)- f\left(x ; W_{(1: i-1)}, \widetilde{W}_{i,q}\right)\right)\right\|_{F} .
\end{aligned}
\end{equation}

The activation functions from this point require zero-in-zero-out (ZIZO; $\sigma(\mathbf{0})=\mathbf{0}$ ) property to be peeled, in addition to the Lipschitzness. Indeed, we can proceed as:

\begin{equation}
\begin{aligned}
    &\left\|\sigma\left(f\left(x ; W_{(1: i)}\right)- f\left(x ; W_{(1: i-1)}, \widetilde{W}_{i,q}\right)\right)\right\|_{F} \\
    \leq &\left\|\left(f\left(x ; W_{(1: i)}\right)- f\left(x ; W_{(1: i-1)}, \widetilde{W}_{i,q}\right)\right)\right\|_{F}
\end{aligned}
\end{equation}

Similar to Eq. \ref{peel i+1}, the i-th layer can be peeled as:
\begin{equation}
    \label{peel i}
    \begin{aligned}
    &\left\|\left(f\left(x ; W_{(1: i)}\right)- f\left(x ; W_{(1: i-1)}, \widetilde{W}_{i,q}\right)\right)\right\|_{F}\\
    =&\left\|\left(W_{i,q}\right) \sigma\left(f\left(x ; W_{(1: i-1)}\right)\right)\right\|_{F} \\
    \leq&\left\|W_{i,q}\right\|_{F} \cdot\left\|\sigma\left(f\left(x ; W_{(1: i-1)}\right)\right)\right\|_{F} \\
    \leq&\left\|W_{i,q}\right\|_{F} \cdot\left\|\left(f\left(x ; W_{(1: i-1)}\right)\right)\right\|_{F}
    \end{aligned}
\end{equation}

To combine Eq. \ref{peel back}, \ref{peel i+1}, \ref{peel i}, we can get the follow inequality:
\begin{equation}
    \begin{aligned}
         sup & \left\|f\left(x ; W_{(1: L)}\right)-f\left(x ; W_{(1: i-1)}, \widetilde{W}_{i,q},\overline{W}_{i+1,q}, W_{(i+2: L)}\right)\right\|_{F} \\
         \leq & \left\|W_{i,q}\right\|_{F} \cdot \left\|\widehat{W}_{i+1,q}\right\|_{F} \cdot \left(\prod_{j \neq i,i+1}\left\|W_{j}\right\|_{F}\right)
    \end{aligned}
\end{equation}
Then we arrive at Eq~\ref{result}. Despite the sub-optimalities from the relaxation, considering the right-hand side of Eq~\ref{result} provides two favorable properties. First, the right-hand side is free of any activation function, and is equivalent to the layerwise distortion. Second, the score can be computed in advance, i.e., does not require re-computing after pruning each filter. In particular, the product term $\prod_{j=1}^{L}\left\|W_{j}\right\|_{F}$ does not affect the pruning decision, and the denominator can be pre-computed with the cumulative sum $\sum_{v \geq u}\left\|W_{l, v}\right\|_{F}^{2} \times\left\|\widehat{W}_{l+1, v}\right\|_{F}^{2}$ for each filter $u$ in $W_{l}$. This computational trick leads us to the SP-LAMP score (Eq. \ref{sp-lamp}).
\begin{equation}
\label{sp-lamp}
\operatorname{score}(u,l):=\frac{\|W_{l,u}\|^{2}_{F} \times \|\widehat{W}_{l+1,u}\|^{2}_{F}}{\sum_{v \geq u}{\|W_{l,v}\|^{2}_{F} \times \|\widehat{W}_{l+1,v}\|^{2}_{F}}}.
\end{equation}

\section{nn-Meter}
\begin{figure}[t]
	\centering
		\centering
		\includegraphics[width=0.32\textwidth]{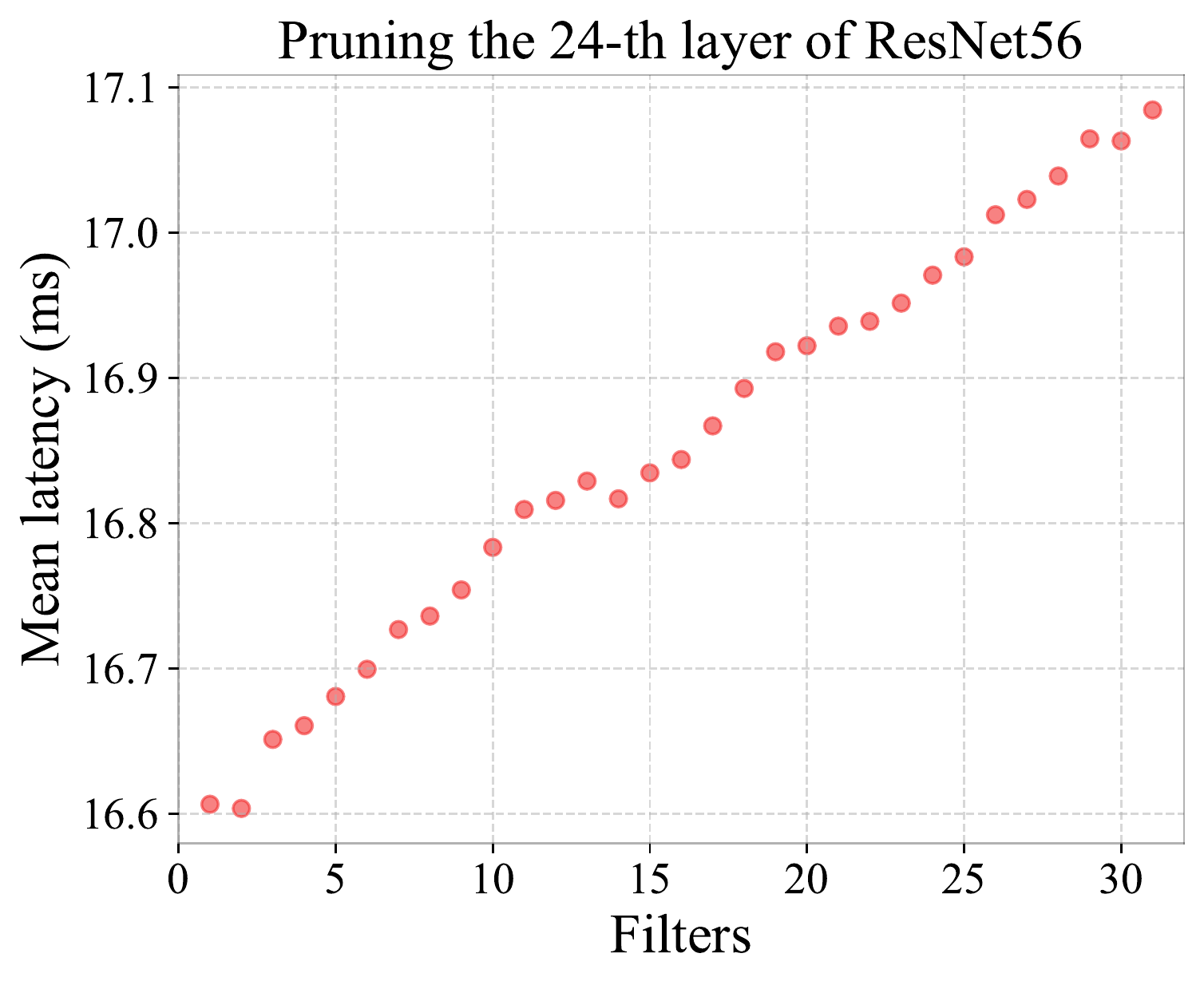}
        \includegraphics[width=0.32\textwidth]{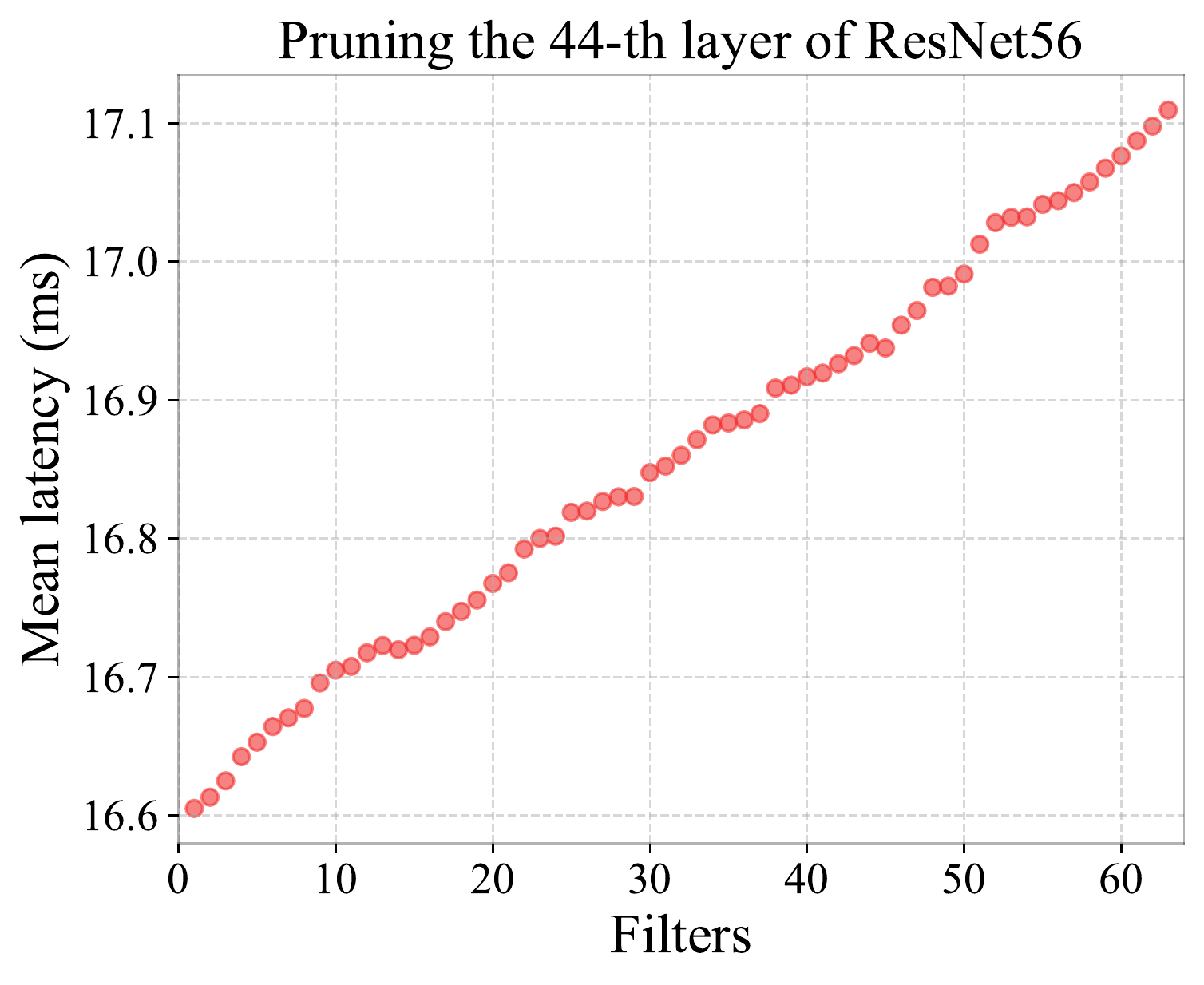}
        \includegraphics[width=0.32\textwidth]{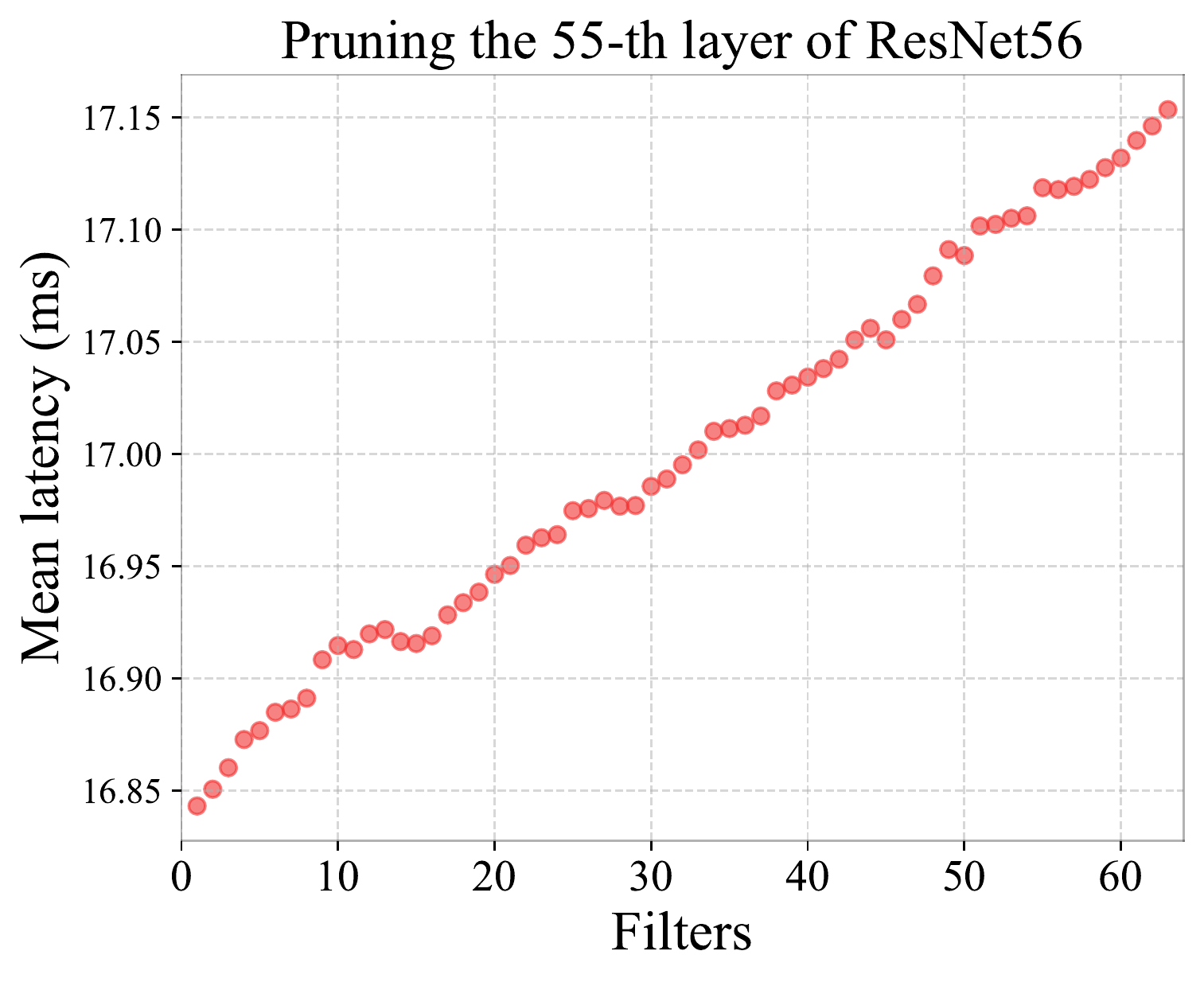}
	\caption{The trend curve of the 24,44,55-th layer of ResNet56 predicted by nn-Meter, every filters removed corresponds to one point in the image.}
	\label{latency_contribution_56}
\end{figure}
\begin{figure}[t]
	\centering
		\centering
		\includegraphics[width=0.32\textwidth]{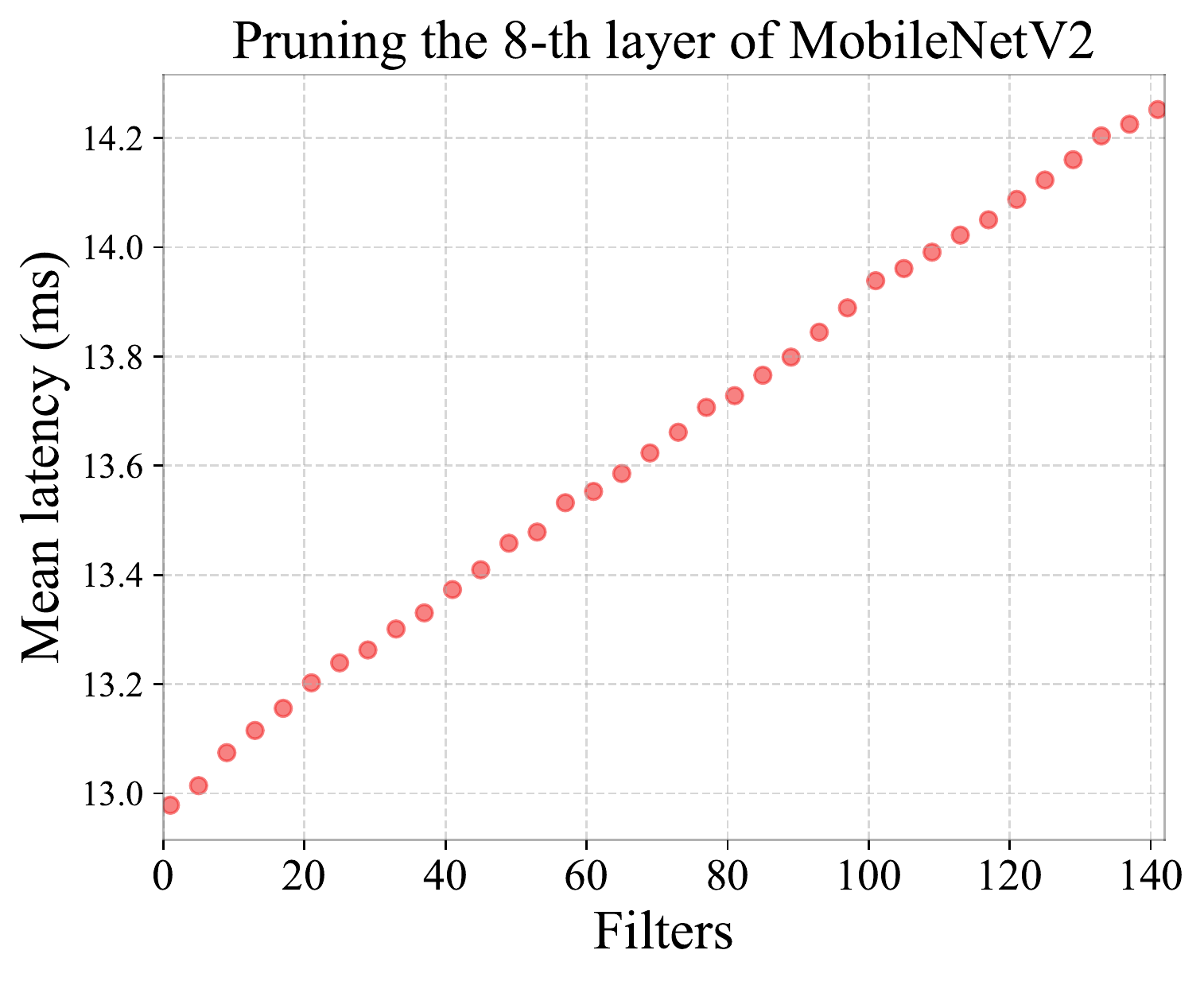}
        \includegraphics[width=0.32\textwidth]{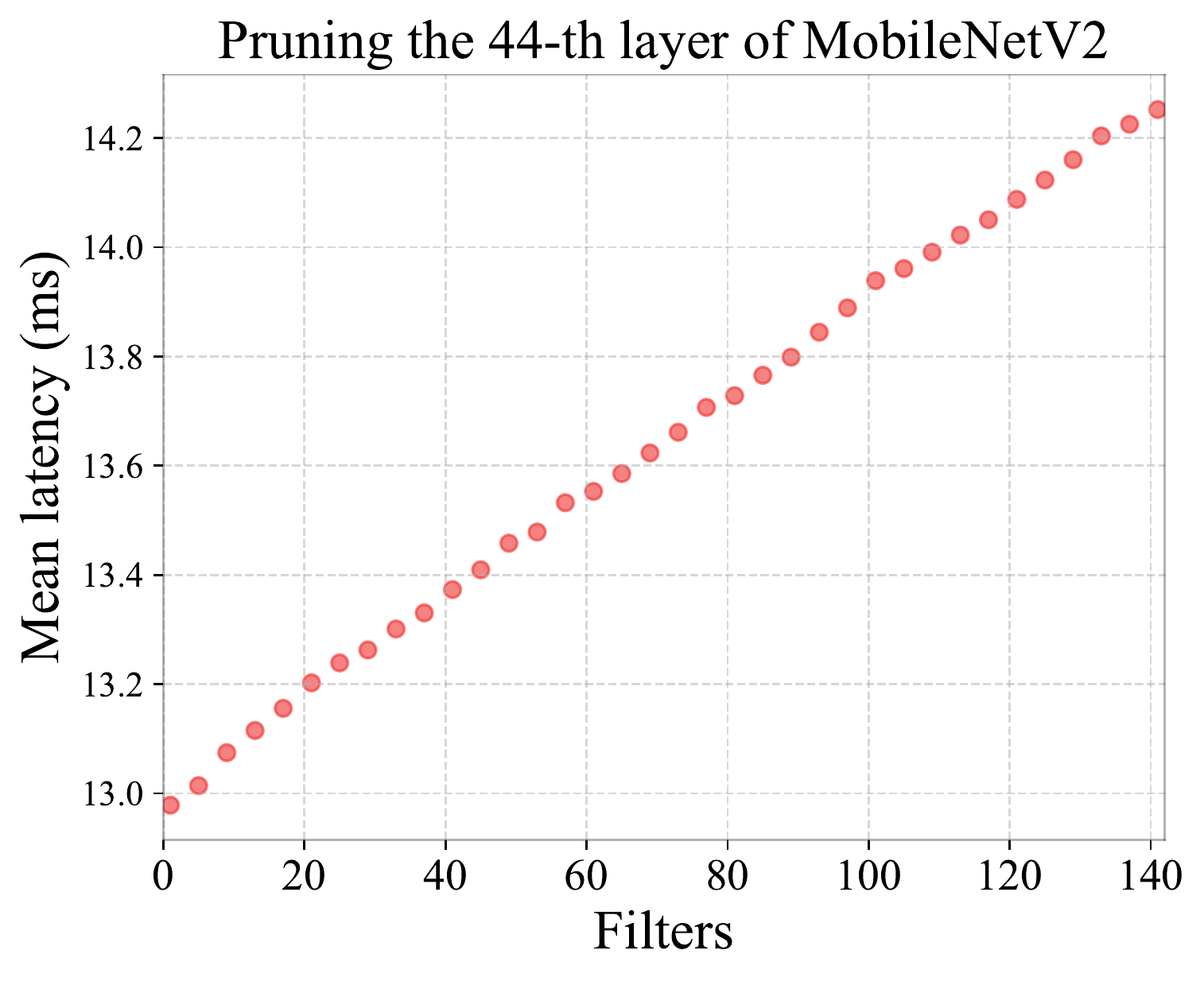}
        \includegraphics[width=0.32\textwidth]{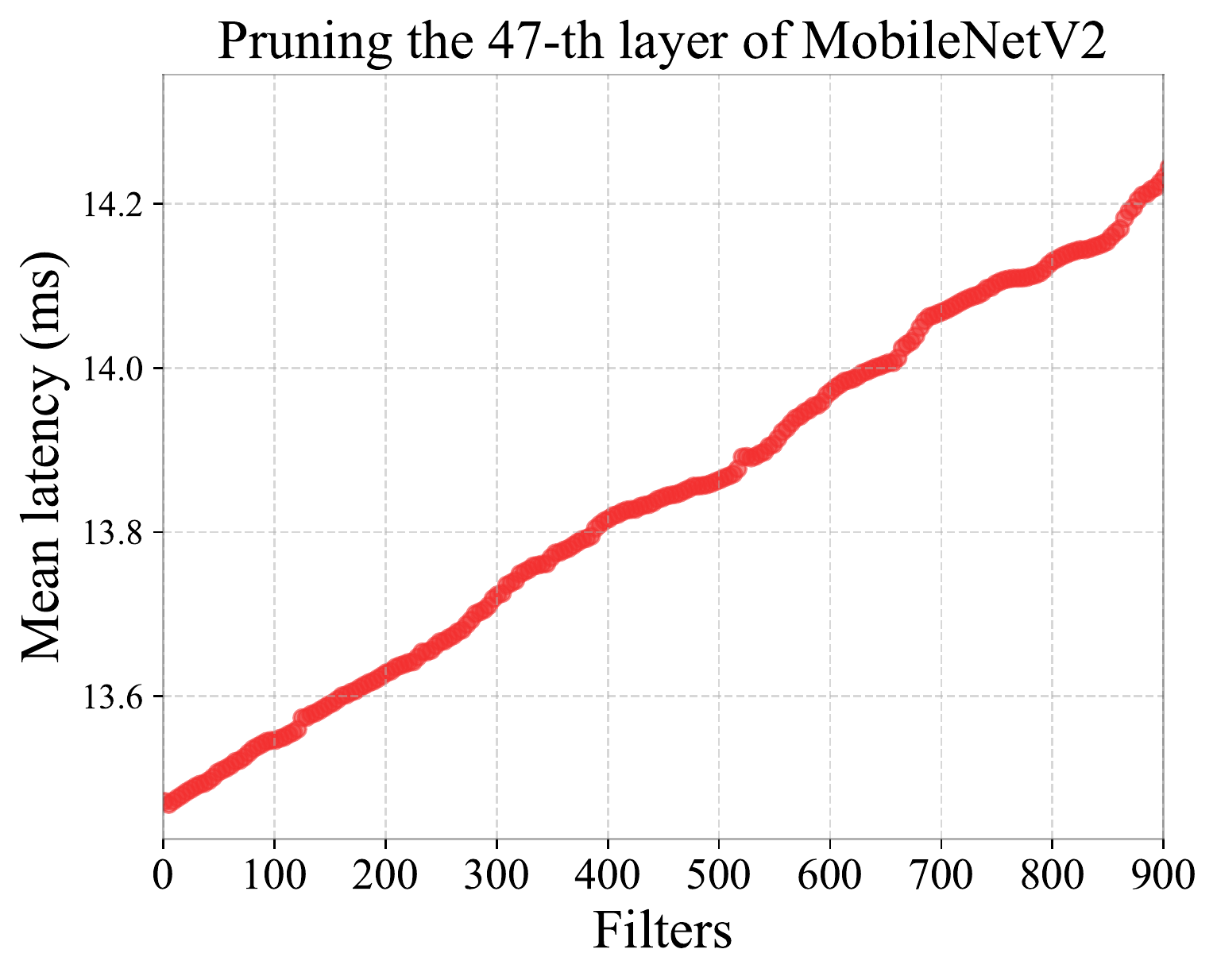}
	\caption{The trend curve of the 8,44,47-th layer of MobileNetV2 predicted by nn-Meter, every filters removed corresponds to one point in the image.}
	\label{latency_contribution_m}
\end{figure}
In our study, we employed the method described in Section 3.2 to construct the latency lookup table for 2080Ti inference. However, this process was time-consuming, taking nearly 2.5 days to complete. To expedite this process, we explored the possibility of utilizing a cost model to streamline the procedure.

nn-Meter\cite{Zhang2021nnMeterTA} is a latency estimation tool designed for deep neural networks (DNNs). By analyzing the structure of a given DNN model, nn-Meter calculates the expected latency for each layer and the entire model on different hardware platforms, including CPUs, GPUs, and mobile devices. It has been widely utilized in research projects such as network compression, acceleration, auto-tuning, and cross-platform deployment, showcasing its effectiveness and practical applicability.

\begin{wrapfigure}{htpb}{5cm}
\centering
\vspace{-7mm}
\includegraphics[width=0.3\textwidth]{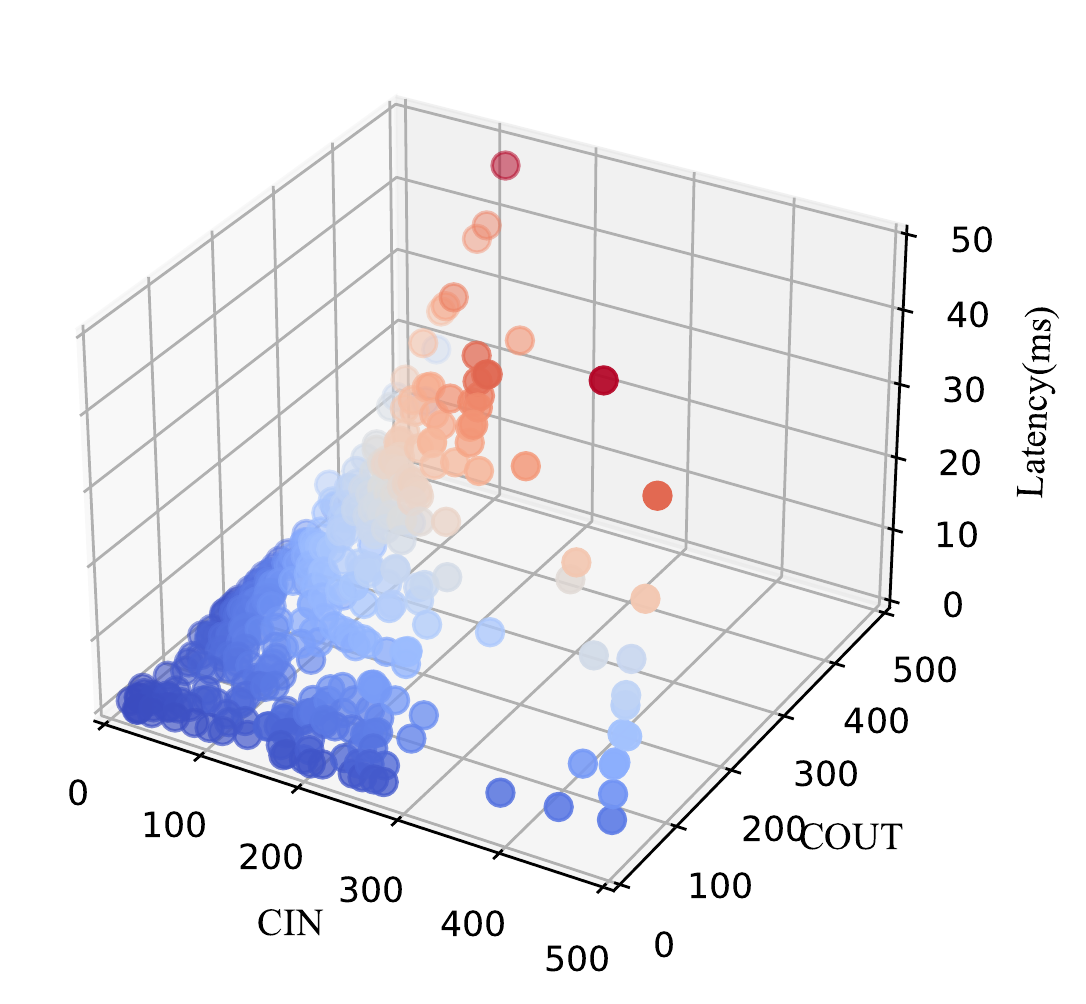}
\caption{Profiled lookup table. (fixed H=W=32, nn-Meter).}
\label{look_up}
\end{wrapfigure}
We utilized nn-Meter to profile the conv-bn-relu kernel using a fixed input size (H = W = 32) and visualized the data in a three-dimensional scatter plot in Figure \ref{look_up}. The plot clearly demonstrates that the latency is influenced by the number of input and output channels, but there is no discernible relationship between them. Interestingly, we observed a purely linear relationship between the variables.

In our work, we employ nn-Meter to collect latency information and quickly obtain profiling data, thereby significantly reducing the time required for profiling.
For ResNet56, we implement the same process as ResNet18. In Figure~\ref{latency_contribution_56}, it is observed that the removal of certain filters leads to a significant jump in latency. We also show the trend curve of MobileNetV2 predicted by nn-Meter in Figure~\ref{latency_contribution_m}.

\section{LAMP}
As our proposed pruning method is an extension of layer-adaptive magnitude-based pruning (LAMP) \cite{lee2020layer} in the direction of structured pruning, we briefly review the idea of LAMP here.

In search of a "go-to" layerwise sparsity for MP, a model-level distortion minimization perspective is token towards magnitude-based pruning (MP). LAMP builds on the observation of Dong \etal \cite{dong2017learning} and Park \etal\cite{park2020lookahead} that each neural network layer can be viewed as an operator, and MP is a choice that incurs minimum $\ell_{2}$ distortion to the operator output (given a worst-case input signal). The perspective further is brang to examine the "model-level" distortion incurred by pruning a layer; preceding layers scale the input signal to the target layer, and succeeding layers scale the output distortion.

Based on the distortion minimization framework, a novel importance score for global pruning, coined LAMP is proposed. The LAMP score is a rescaled weight magnitude, approximating the model-level distortion from pruning. Importantly, the LAMP score is designed to approximate the distortion on the model being pruned, i.e., all connections with a smaller LAMP score than the target weight is already pruned. Global pruning  with the LAMP score is equivalent to the MP with an automatically determined layerwise sparsity. At the same time, pruning with LAMP keeps the benefits of MP intact; the LAMP score is efficiently computable, hyperparameter-free, and does not rely on any model-specific knowledge.

Consider a fully connected neural network with depth $d$ and weight tensors $W^{(1)}, . . . , W^{(d)}$ (convolutional neural networks have the same conclusion \cite{sedghi2018singular}). Let $x$ be an input vector whose output is $ f(x;W^{(1:d)})=W^{(d)}\sigma(W^{(d-1)}\sigma(...W^{(2)}\sigma(W^{(1)}x)...) $. For the weight tensor $W^{(i)}\in \mathbb{R}^{m \times n}$ of the i-th fully connected layer, the pruned version of $W^{(i)}$ is represented as $\widetilde{W}^{(i)}:= M^{(i)} \times W^{(i)}$, where $M^{(i)}$ is an $m \times n$ binary matrix with elements only 0s and 1s, and $||M^{(i)}||_0 \leq \kappa_i$. Let $\xi_i \in \mathbb{R}^n$ denote the input of the i-th layer, the optimization objective of the current layer in LAMP is: $\min _{\|M^{(i)}\|_0 \leq \kappa_i} \sup _{\|\xi_i\|_2 \leq 1}{\|W^{(i)} \xi_i-(M^{(i)} \odot W^{(i)}) \xi_i\|_{2}}$, and the resultant problem of the entire network can be described as follows:
\begin{equation}
\min _{\sum_{i=1}^{d}\left\|M^{(i)}\right\|_{0} \leq \kappa} \sup _{\|x\|_{2} \leq 1}\left\|f\left(x ; W^{(1: d)}\right)-f\left(x ; \widetilde{W}^{(1: d)}\right)\right\|_{2}
\end{equation}
where $\kappa$ denotes the model-level sparsity constraint. 
Under the greedy strategy, the connection with the lowest importance score is removed every iteration. By the definition of the spectral norm and the condition of $\|A\| \leq \|A\|_F$, the above optimization can be relaxed by using the following upper bound of the model output distortion: 
\begin{equation}
\begin{aligned}
   \sup _{\|x\|_{2} \leq 1}\left\|f\left(x ; W^{(1: d)}\right)-f\left(x ; W^{(1: i-1)}, \widetilde{W}^{(i)}, W^{(i+1: d)}\right)\right\|_{2} \\
   \leq  \frac{\left\|W^{(i)}-\widetilde{W}^{(i)}\right\|_{F}}{\left\|W^{(i)}\right\|_{F}}\left(\prod_{j=1}^{d}\left\|W^{(j)}\right\|_{F}\right)
\end{aligned}
\end{equation}

Flattened weight tensor $W^{(i)}$ to a one-dimensional vector and sort it to satisfy $|W[u]| \leq |W[v]|$ whenever $u < v$. Since the product term $\prod_{j=1}^{d}\left\|W^{(j)}\right\|_{F}$ is a constant value, then the importance score for the u-th index of the weight tensor $W$ is defined as:
The LAMP score for the $u$-th index of the weight tensor $W$ is then defined as:
\begin{equation}
\label{lamp}
\operatorname{score}(u ; W):=\frac{(W[u])^{2}}{\sum_{v \geq u}(W[v])^{2}}
\end{equation}

Informally, the LAMP score (Eq.~\ref{lamp}) measures the relative importance of the target connection among all surviving connections belonging to the same layer, where the connections with a smaller weight magnitude (in the same layer) have already been pruned. As a consequence, two connections with identical weight magnitudes have different LAMP scores, depending on the index map being used. Once the LAMP score is computed, we globally prune the connections with smallest LAMP scores until the desired global sparsity constraint is met; the procedure is equivalent to performing MP with an automatically selected layerwise sparsity. 

\section{HALP}
Hardware-aware Latency Pruning (HALP) \cite{shen2021halp} addresses the problem of latency-constrained structural pruning using an augmented Knapsack algorithm. However, the proposed knapsack solver in HALP may result in the elimination of all neurons in a layer. Furthermore, HALP considers the latency of each neuron solely based on the current layer, without taking into account skip connections and the potential impact of the current layer on other layers. In our algorithm, we improve the latency acquisition strategy and employ a grouping knapsack solver as an alternative to the original 0-1 knapsack solver. Additionally, HALP performs pruning on batch normalization layers, and the importance of the n-th neuron in the l-th layer is computed as follows:
\begin{equation}
\mathcal{I}_{l}^{n}=\left|g_{\gamma_{l}^{n}} \gamma_{l}^{n}+g_{\beta_{l}^{n}} \beta_{l}^{n}\right|,
\end{equation}

where $g$ denotes the gradient of the weight, $\gamma_{l}^{n}$ and $\beta_{l}^{n}$ are the corresponding weight and bias from the batch normalization layer, respectively. This important score is approximated from the importance of neurons using the Taylor expansion of the loss change\cite{molchanov2019importance}. The intuition is that if weights are changed little from their initial random values during the network’s learning process, then they may not be too important. This method would be identical to sparsification techniques based on absolute magnitude if we consider the change with respect to a (contrived) starting state of all-zero weights. However, when we remove the entire filter, it will actually cause a large loss to the neural network, and with the increase of the pruning ratio, a smaller network is more sensitive to sparsity. This assumption is untenable. 
\begin{table}[h]

\caption{Comparison of diffenrent pruning methods. }
\centering
\begin{tabular}{ccccccc}
\toprule
\multirow{2}{*}{Method} & Structured & Importance& Latency & Consider & Minimum& Knapsack\\ & Pruning & Score& Guided & Connection & One Filter & Solver\\
                         \midrule
LAMP\cite{lee2020layer} & \ding{55} & LAMP & \ding{55} & \ding{55} & \ding{51} & - \\
HALP\cite{shen2021halp} & \ding{51} & $l$2-norm& \ding{51} & \ding{55}& \ding{55}& 0-1 knapsack \\
Ours & \ding{51} & SP-LAMP & \ding{51} & \ding{51} & \ding{51}& group knapsack \\

\bottomrule
\end{tabular}

\label{difference}

\end{table}
To address this problem, we propose SP(Structured Pruning)-LAMP, by deriving a global importance score LAMP\cite{lee2020layer} from unstructured pruning to structured pruning. 
In Table~\ref{difference}, we compare our method with HALP and the unstructured pruning method LAMP, highlighting the differences.

\section{Implementation details of experiments}
\textbf{Convert latency in float to int.} Solving the problem of selecting filters requires our proposed grouping knapsack algorithm, which requires latency contributions and latency constraints to be numbers of type int, To convert the measured latency from
a full precision floating-point number to integer type, we multiply the latency (s) by 1000000 and perform rounding. Accordingly, we also scale and round the latency constraint value. The latency predicted by the nn-Meter is in milliseconds, we multiply the latency by 10000.

\textbf{Pre-processing.} CIFAR-10 dataset is augmented with random crops with a padding of 4 and random horizontal flips. We normalize both training and test datasets with constants
\begin{center}
$(0.4914, 0.4822, 0.4465), (0.2023, 0.1994, 0.2010)$ 
\end{center}

\textbf{Pruning process.} We use ten stages to remove the total latency ratio respectively. Each stage first uses the mask and then fine-tunes the pruned model to restore the accuracy, and then proceeds to the next knapsack algorithm to select the pruned filter. With optimizer of SGD, the learning rate is set to 0.01, the weight decay is set to 0.0005, and epochs is set to 200, the pruning process is performed on 2080Ti, and the hard prune is performed after the pruning fine-tuning is completed.

\end{document}